\documentclass{article}
\usepackage[T1]{fontenc}

\usepackage{iclr2027_conference,times}
\usepackage{booktabs}
\usepackage{amsmath}
\usepackage{amssymb}
\usepackage{graphicx}
\usepackage{xcolor}
\usepackage{xspace}
\usepackage{multirow}
\usepackage{array}
\usepackage{tabularx}

\newcommand{\trace}{\textsc{TRACE}\xspace}
\newcolumntype{Y}{>{\centering\arraybackslash}X}
\newcolumntype{L}{>{\raggedright\arraybackslash}X}

\graphicspath{{figures/}{./}}

\title{TRACE: Target-State Risk-Aligned\\Conditional Execution for\\Embodied Tracking}

\author{\hspace*{-\tabcolsep}\parbox[t]{\textwidth}{\centering
\setlength{\parskip}{0pt}
{\normalfont\fontsize{11}{16}\selectfont\bfseries
Mingyang Lyu$^{1,2,5,*}$\hspace{1.2em}Gongce Cao$^{2,*}$\hspace{1.2em}Yiyang Jia$^{2}$\hspace{1.2em}Lu Jia$^{4}$ \\[1pt]
Erliang Lin$^{1}$\hspace{1.2em}Yuhao Zhang$^{3}$\hspace{1.2em}Yinqian Sun$^{1,*}$ \\[1pt]
Pin Lyu$^{1,\dagger}$\hspace{1.2em}Feifei Zhao$^{1,\dagger}$\hspace{1.2em}Bo Wang$^{1,\dagger}$\endgraf}
\vspace{8pt}
{\normalfont\fontsize{9}{12}\selectfont $^{1}$ Institute of Automation, Chinese Academy of Sciences \\
$^{2}$ University of the Chinese Academy of Sciences \\
$^{3}$ Xi'an Jiaotong University\qquad $^{4}$ University of Science and Technology Beijing \\
$^{5}$ Zhongguancun Academy\endgraf}
\vspace{6pt}
{\normalfont\fontsize{8.5}{10}\selectfont $^*$ Equal contribution.\qquad $^{\dagger}$ Corresponding authors.\endgraf}
}\hspace*{-\tabcolsep}}
\iclrfinalcopy

\begin{document}
\maketitle

\begin{abstract}
	Embodied visual tracking requires a robot to follow a language-specified person under
	occlusion, look-alike distractors, and ambiguous instructions. We show that improving
	tracking persistence can simultaneously increase safety violations. Using a compact
	$0.6$B vision-language-action (VLA) tracker with a frozen backbone, we update only the
	planner head with proximal policy optimization (PPO). PPO improves success and following
	rates across all three EVT-Bench settings, but increases the collision rate on
	distracted tracking because episodes that previously ended in target loss can instead
	end in contact. We introduce \trace{}, an execution-time target-state risk gate that
	preserves the learned policy and intervenes selectively. A lightweight risk head
	processes $84$ cues, including candidate boxes, identity margins, a learned
	range-and-bearing probe, and the policy's commands. When the calibrated risk exceeds a
	threshold, \trace{} rescales the translational component of the next waypoint while
	leaving yaw unchanged. The executed action is identical to PPO's on more than $98\%$ of
	steps. In paired evaluations on identical episodes, \trace{} reduces the
	distracted-tracking collision rate by $3.96$ percentage points, while further improving
	success and following rates. Matched slowdown controls without target-state timing do
	not reproduce this improvement. Concurrent evaluation reproduces a collision reduction
	of $3.77$ points, compared with $0.29$ points for identical-arm null pairs, indicating
	that the effect exceeds evaluation variability. These results show that
	target-state-aware execution can recover the safety cost of policy post-training with
	minimal intervention.
\end{abstract}

\section{Introduction}
\label{sec:intro}

\begin{figure}[t]
	\centering
	\includegraphics[width=0.96\linewidth]{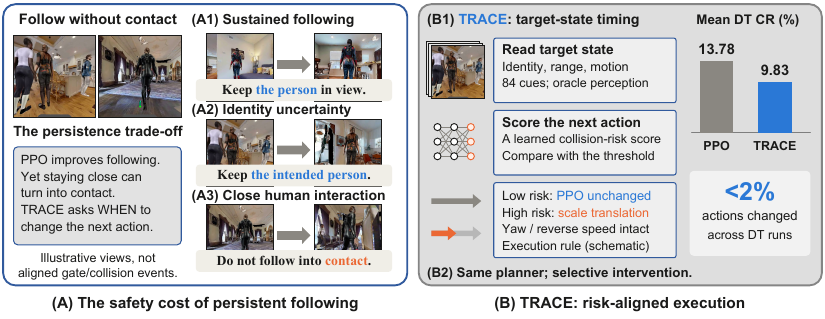}
	\caption{\textbf{The safety cost of persistent following, and the gate that removes
		it.} \textbf{(A)} The robot must keep a language-specified person in view under
		distractors and ambiguity; PPO on the planner head makes it persist longer, and on
		distracted tracking part of that persistence ends in contact. \textbf{(B)} \trace{}
		scores the next action from target-state cues and rescales its translation only
		when the score crosses a threshold; the action is unchanged on over $98\%$ of steps.}
	\label{fig:teaser}
\end{figure}

Embodied visual tracking (EVT) requires an agent to continuously follow a
language-specified person from egocentric observations over long
horizons~\citep{luo2018sarl,zhong2024evt,zhang2025trackvla}. The task requires the joint
solution of target identification, temporal association, and motion planning under
occlusion, look-alike distractors, and ambiguous instructions. Recent
vision-language-action (VLA) models integrate visual perception, language
understanding, and robot control~\citep{brohan2022rt1,brohan2023rt2,kim2024openvla,black2024pi0}.
The strongest reported results on EVT-Bench come from a $7$B-scale model trained on large-scale data~\citep{zhang2025trackvla}, whereas deployment favors compact models with limited adaptation cost and low execution overhead.

We construct a compact three-stage tracking system based on a frozen Qwen3-0.6B
backbone. The supervised base model keeps the vision encoder, language backbone, and
recognition adapter fixed, while a planner head predicts eight future waypoints. We then apply PPO~\citep{schulman2017ppo,schulman2015gae} only to the planner head, which improves success and following rates across all three EVT-Bench settings: the planner head is an effective control point for tracking persistence, and the representation backbone need not be updated. The improvement carries a safety cost in distracted tracking. Relative to the supervised model, PPO reduces early target loss but increases collisions, changing the terminal outcome of some episodes from target loss to physical contact (Figure~\ref{fig:teaser}A).

This observation exposes a limitation of optimizing persistence with a scalar task
reward. Existing safe-control and recovery
methods~\citep{achiam2017cpo,ray2019safetygym,johannink2019residual,thananjeyan2021recoveryrl,ames2019cbf,dalal2018safe}
typically impose geometric constraints or respond to local collision conditions. They
do not explicitly model the target-state uncertainty that precedes collisions during
distracted tracking. In our experiments, a distance-only execution filter does not
recover the safety degradation introduced by PPO (\S\ref{sec:why_not_reward}). These
results motivate an execution-time mechanism that uses the tracker's target-state
estimate to intervene selectively while preserving the persistence learned by PPO.

We therefore introduce \trace{}, a target-state risk gate applied after policy
prediction (Figure~\ref{fig:teaser}B). A lightweight head receives $84$ target-related cues, including candidate
bounding boxes, identity margins, a learned range-and-bearing probe, and the policy's
commands and predicted waypoints. The head produces a collision-risk score. Once the
score exceeds a calibrated threshold, \trace{} rescales the translational component of
the next waypoint while preserving its yaw component. At all other steps, the executed
action is identical to the PPO action. The gate reduces the distracted-tracking
collision rate (CR) from $13.78$ to $9.83$, below the supervised baseline, while further improving success and following rates (SR, FR).

Our contributions are as follows.
\vspace{-0.35em}
\begin{itemize}
	\setlength{\leftmargin}{1.15em}
	\setlength{\labelsep}{0.35em}
	\setlength{\itemsep}{0pt}
	\setlength{\parsep}{0pt}
	\setlength{\topsep}{0pt}
	\item \textbf{A compact VLA tracking stack with an empirical analysis of post-training
	safety costs.} We develop a $0.6$B frozen-backbone VLA tracker and restrict PPO
	post-training to its planner head. This design improves success and following rates
	on all three EVT-Bench settings but raises the distracted-tracking collision rate from
	$11.01$ to $13.78$, establishing a measurable trade-off between persistence and
	collision safety under planner-only post-training.
	\item \textbf{\trace{}: selective target-state risk gating for safe execution.} We
	introduce an execution-time gate that predicts collision risk from target-state cues
	and modifies only the translational component of the next waypoint when the risk
	exceeds a calibrated threshold. On distracted tracking, \trace{} reduces the collision
	rate by $3.96$ percentage points while increasing the success rate by $2.03$ points and
	the following rate by $2.14$ points, at approximately $50\,\mu$s per step and fewer than
	$2\%$ of executed actions changed. Matched slowdown controls without target-state timing
	remain at the PPO reference, and the effect persists across two scene corpora, three
	collision geometries, held-out episodes, and a later code version.
	\item \textbf{Concurrent evaluation for separating intervention effects from benchmark
	variability.} Re-evaluating an unchanged policy shifts collision rate by up to $1.4$
	points and success rate by up to $2.7$, so we evaluate the gated and ungated policies
	concurrently together with identical-arm null pairs. Under this protocol, \trace{} produces a collision-rate
	change of $-3.77$ points, compared with $-0.29$ points for identical-arm pairs,
	providing a paired estimate of the intervention effect relative to evaluator
	variability.
\end{itemize}
\vspace{-0.45em}

\section{Related Work}
\label{sec:related}

\paragraph{Embodied visual tracking.}
Early work casts the task as reinforcement learning over egocentric
observations~\citep{luo2018sarl,luo2019rl}, later hardened against evasive targets and
occlusion by adversarial and structured
variants~\citep{zhong2019advat,zhong2021ts,zhong2023rspt}. A second line decouples the
problem into a visual tracker and a controller, pairing detection or re-identification
with image-based servoing~\citep{gupta2016ibvs,bhat2019dimp,maalouf2024followanything}.
TrackVLA~\citep{zhang2025trackvla} instead unifies recognition and planning in a single
$7$B model and releases EVT-Bench, and TrackVLA++~\citep{liu2025trackvlapp} extends it with explicit reasoning and memory. Both train large backbones end-to-end for the task. In
contrast, we hold a compact frozen-backbone tracker fixed and ask what post-training its
planner head alone does to collision behavior. Appendix~\ref{app:related} covers the
adjacent literature on robot following, social navigation, and pursuit--evasion.

\paragraph{Post-training and runtime safety.}
Post-training a supervised policy with on-policy RL is standard practice for language,
vision-language, and embodied controllers~\citep{brohan2023rt2,kim2024openvla,black2024pi0},
with PPO and generalized advantage estimation the usual
choice~\citep{schulman2017ppo,schulman2015gae}. Safety is typically added as a separate
channel that edits or constrains a proposed action: control-barrier
functions~\citep{ames2019cbf}, learned safety layers~\citep{dalal2018safe}, recovery
policies~\citep{thananjeyan2021recoveryrl}, residual
corrections~\citep{johannink2019residual}, and constrained MDPs that fold the constraint
into the objective~\citep{achiam2017cpo,ray2019safetygym}. Each of these reads geometry,
known dynamics, or a constraint signal defined independently of the followed person. In
contrast, we condition the intervention on the tracker's own estimate of the target's
state, which is the quantity that changes when persistent following turns into contact.

\paragraph{Pretrained components and simulation benchmarks.}
The quantities our risk head reads are produced by components we do not retrain:
re-identification confidence from tracking-by-detection
pipelines~\citep{wojke2017deepsort,zhang2022bytetrack}, frozen vision-language
backbones~\citep{radford2021clip,oquab2023dinov2,bai2025qwen25vl}, and waypoint heads of
the kind used in diffusion and transformer
policies~\citep{ho2020ddpm,chi2023diffusionpolicy,zhao2023act}. EVT-Bench itself is built
on Habitat~\citep{savva2019habitat,puig2023habitat3} with HM3D and Matterport3D
scenes~\citep{ramakrishnan2021hm3d,chang2017matterport3d}, and inherits its episode and
instruction conventions from vision-language
navigation~\citep{anderson2018vln,krantz2020vlnce}. We take all of these as given;
Appendix~\ref{app:related} lists the remaining references.

\section{Method}
\label{sec:method}

\begin{figure}[t]
	\centering
	\includegraphics[width=0.9\linewidth]{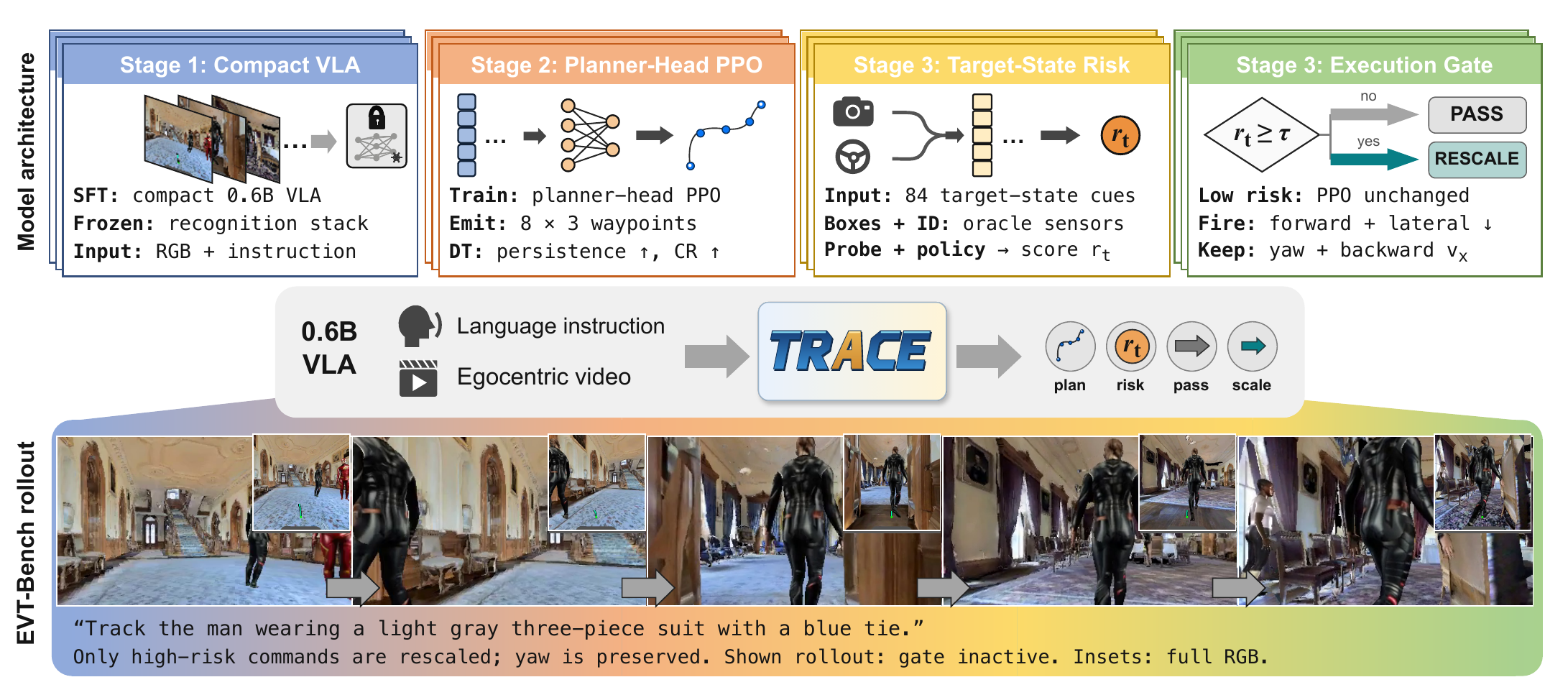}
	\caption{\textbf{The tracking stack.} Supervised tracker, then PPO on the planner
		head alone, then a risk score $r_t$ that gates execution: below $\tau$ the PPO
		command passes through; above it, forward and lateral motion are rescaled and yaw is
		preserved. The $84$ cues use EVT-Bench's own detection and identity
		(\S\ref{sec:deployability}). The rollout below is one episode on which the gate
		never fires.}
	\label{fig:main_overview}
\end{figure}

\paragraph{Problem and overview.}
At each step $t$, the agent receives a language instruction $I$ that
specifies the target and an egocentric RGB history $O_t$. A compact VLA
tracker maps $(I,O_t)$ to an $8{\times}3$ waypoint sequence
$W_t=\{w_{1,t},\ldots,w_{8,t}\}$, where each waypoint is
$w_{i,t}=(v_x,v_y,\omega)$. Only the first waypoint is executed through
a PD controller, after which the agent observes again and replans. An
episode is successful when the agent keeps the described target within
the evaluator's following band without terminating in collision. Our
goal is to increase persistence without converting lost-target failures
into collisions. The construction has three components: a compact supervised tracker
whose recognition components are then frozen (\S\ref{sec:tracker}), PPO on its planner
head alone (\S\ref{sec:ppo}), and a gate that rescales the next waypoint only when the
risk score is high (\S\ref{sec:risk_shield}; Figure~\ref{fig:main_overview}). Each component
leaves the previous one unchanged, so we can measure the effect of each separately.

\subsection{A Compact VLA Tracker}
\label{sec:tracker}

The base tracker couples a vision encoder, a Qwen3-0.6B language backbone, and a
recognition adapter that projects visual tokens into the language space, followed by a
planner head that emits the $8\times3$ waypoint sequence. It is trained by supervised fine-tuning (SFT; Appendix~\ref{app:training}) and then serves as the fixed reference: the encoder, backbone, and adapter are never updated again, so every later change is attributable to the planner head or the gate.
The limitation is persistence rather than recognition: most failures are episodes in which the target is dropped early and never recovered, rather than episodes in which the wrong person is followed.

\subsection{Planner-Head PPO}
\label{sec:ppo}

To improve persistence without disturbing recognition, we freeze the vision
encoder, language backbone, and recognition adapter; only the deterministic waypoint head $\mu_\theta$, an exploration term, and a value head are updated. Training samples actions around the waypoint mean, $\hat{\mathbf{w}}_t \sim \mathcal{N}(\mu_\theta(h_t), \mathrm{diag}(\sigma^2))$, and evaluation uses the mean itself. All RL-style variants share the same reward:
\begin{equation}
	\label{eq:reward}
	r_t = w_f\,\mathbb{1}[\mathrm{follow}_t]
	- w_c\,\mathbb{1}[\mathrm{collide}_t]
	+ w_d\,(d_{t-1}-d_t)
	- w_{\mathrm{far}}\,\mathbb{1}[d_t>4\,\mathrm{m}],
\end{equation}
where $(w_f,w_c,w_d,w_{\mathrm{far}})=(1.0,5.0,0.5,0.3)$. The follow and collision indicators come from the EVT-Bench following band and simulator contact flag, and $d_t$ is the geodesic distance to the target. We optimize this objective with PPO and GAE~\citep{schulman2017ppo,schulman2015gae} over
two rounds of $200$ on-policy episodes. Because the perception path is frozen, this
update cannot change which person the tracker believes it is following; it can only change how long the policy commits to that belief, so the collision term in Eq.~\ref{eq:reward} does not prevent the failure it is meant to penalize. The update
improves following but does not by itself indicate when the next action is unsafe.

\paragraph{Single-target tracking.}
\label{sec:stt_method}\label{sec:stt_selection}
When the target is unambiguous, the risk score rarely crosses a threshold that changes the trajectory without lowering SR, so we treat this setting separately. Starting from the collision-anchored planner, we run one further low-learning-rate PPO round at $c{=}16$ and $c{=}18$, with no gate or recovery residual. For each training run a selection subset picks the penalty level with lower CR, and the resulting mix is evaluated on an independent subset and on the full five-run setting; the selection is stable under a wider grid (Appendix~\ref{app:stt_attempts}). The single-target tracking (STT) result is therefore planner-side; distracted tracking (DT) and ambiguity tracking (AT) test conditional execution at runtime.

\subsection{Target-State Risk Gate}
\label{sec:risk_shield}

The same waypoint can be safe when the target is clearly visible and dangerous when
identity, range, or recent motion indicate a close and ambiguous situation. We therefore assemble a target-state vector $x_t$ of $84$ quantities, all computed online
before the step executes (Appendix~\ref{app:feature_table}): detector-side box geometry
and confidence together with the top-2 identity margin, distance and bearing bins from a
learned grounding probe, memory age and status flags, the previous command and the
planner's predicted waypoints, image-plane flags, a visible-human count, and the step
index. None of these reads simulator pose, geodesic distance, or the benchmark's
following bands; those privileged quantities are excluded by construction, so the score
is computable from what the deployed tracker already holds. In this benchmark, however,
the candidate boxes and the target association are provided by the benchmark's own humanoid sensors rather than by a detector we train (\S\ref{sec:deployability}).

A risk head $f_\phi$ maps $x_t$ to a collision-risk score $r_t = f_\phi(x_t)\in[0,1]$: a small feed-forward network of $28$k parameters with auxiliary lost-target and failure outputs (Appendix~\ref{app:heads}). It is
fitted offline on distracted-tracking rollouts of earlier policies, with a positive label
on the steps inside a fixed window preceding a collision termination, so the head is
trained to anticipate contact rather than to report the current distance. The score ranks risk well but is not a calibrated probability, so $\tau$ is set from the trigger rate and the collision reduction rather than from a nominal probability level (Appendix~\ref{app:calibration_detail}). At inference
time the proposed waypoint is executed as
\begin{equation}
	\label{eq:risk_shield}
	w^{\mathrm{exec}}_{1,t} =
	\begin{cases}
		\bigl(\min(v_x,0) + s_f\max(v_x,0),\; s_l v_y,\; s_\omega \omega\bigr), &
		r_t \ge \tau,                                                   \\[2pt]
		w^{\mathrm{base}}_{1,t},                 & \text{otherwise.}
	\end{cases}
\end{equation}
Whenever $r_t<\tau$, the deployed policy is identical to planner-head PPO; when the gate fires, only the current action is rescaled: yaw is kept ($s_\omega=1$), lateral motion is attenuated, and forward motion is attenuated only when the command is forward, because slowing a retreat does not reduce contact. The gate therefore cannot introduce a motion the planner did not propose. The DT operating point is $(\tau, s_f, s_l, s_\omega)=(0.868,0.25,0.60,1.00)$, calibrated on diagnostic trajectories; the score is learned rather than certified, so a high-risk step the head misses is executed unchanged. Because only the executed waypoint is edited and no policy parameter is
touched, this operating point can be re-calibrated for a new setting without repeating
the post-training. For AT we fit a separate risk head on AT rollouts and calibrate its own threshold and
decision rule (Appendix~\ref{app:calibration}); the two operating points are denoted jointly as the gate.
This design makes the mechanism testable, because the amount of attenuation can be held
fixed while its timing is varied. The controls we report do exactly that: the same
attenuation applied at every step, a random trigger matched to the gate's measured firing
rate, and the gate's own score delayed in time so that it fires as often but at the wrong
steps. If timing aligned with the target state does not matter, any of them should
reproduce the gate's safety and success numbers.

\section{Experiments}
\label{sec:results}

\subsection{Setup}
\label{sec:setups}

\textbf{Benchmark.}
We evaluate on EVT-Bench~\citep{zhang2025trackvla}. Single-target tracking (STT) has
one language-specified target in a clean scene; distracted tracking (DT) adds agents
with similar appearance and motion; ambiguity tracking (AT) relaxes the instruction so
the policy must disambiguate the intended target. Each setting contributes $1{,}380$
evaluated episodes per run (Appendix~\ref{app:eval_protocol}), and episodes terminate on
sustained following, collision, or step-budget exhaustion.

\paragraph{Baselines.}
Planner-head PPO (PPO below) is initialized from the supervised model and the gate is deployed on the resulting policy, so internal comparisons isolate post-training from the execution-time intervention. Panel~(a) of Table~\ref{tab:main_compact} lists four prior trackers with the numbers published by TrackVLA, whose trained models are not available: IBVS~\citep{gupta2016ibvs,liu2023groundingdino}, EVT~\citep{zhong2024evt,openai2024gpt4o,yang2023som}, and the $7$B Uni-NaVid~\citep{zhang2024uninavid} and TrackVLA~\citep{zhang2025trackvla}.

\paragraph{Metrics.}
We report success rate (SR), following rate (FR), and collision rate (CR) in percent, following the EVT-Bench protocol~\citep{zhang2025trackvla}. FR is the pooled fraction of episode steps in the following band (the official aggregator's ratio), and CR is the fraction of episodes that end in collision. All compact-VLA numbers are obtained with the corrected full-episode evaluator (Appendix~\ref{app:eval_protocol}); five-run results use five planner-head training runs evaluated under the same configuration.

\paragraph{Evaluation protocol.}
\label{sec:protocol_scope}
EVT-Bench's released evaluator leaks episode state across resets; correcting it shifts DT and AT metrics by several points, and all our rows use the corrected evaluator on the same $1{,}380$ episodes (Appendix~\ref{app:eval_protocol}). Even so, repeated evaluations of one unchanged
policy differ by up to $1.4$ CR and $2.7$ SR points. Every \trace{}-versus-PPO delta is therefore paired on the same episodes and the same five training runs, with episode-level bootstrap intervals and a sign-flip test on collisions (Table~\ref{tab:paired_stats}); the primary DT comparison is additionally evaluated with both arms concurrent and with null pairs whose arms are identical.

\subsection{Main Results}
\label{sec:main_results}

Planner-head post-training and the gate contribute separately (Table~\ref{tab:main_compact}; panel~(a) gives published numbers as context under different protocols, panel~(b) is ours under one evaluator, the same episodes, and five training runs). Planner-head PPO raises SR and FR over the SFT base on every setting ($+8.19/+5.18$ on STT, $+2.61/+0.98$ on DT, $+4.54/+2.66$ on AT). On DT it also raises CR from $11.01$ to $13.78$. Adding the gate removes this collision cost without giving back the persistence: DT CR falls to $9.83$, below the SFT base, SR and FR rise further, and the run-to-run spread of CR narrows from $\pm1.31$ to $\pm0.49$. On STT, where the risk score never crosses $\tau$, per-run model selection adds $+1.06$ SR and $+0.92$ FR at unchanged CR (\S\ref{sec:stt_selection}). On AT the gated row is the AT-calibrated gate.

\begin{table*}[htbp]
	\centering
	\footnotesize
	\setlength{\tabcolsep}{3.2pt}
	\renewcommand{\arraystretch}{1.10}
	\resizebox{\textwidth}{!}{%
		\begin{tabular}{@{}lccc@{}}
			\toprule
			\textbf{Method} & \textbf{STT} & \textbf{DT} & \textbf{AT} \\
			                & \textbf{SR$\uparrow$ / FR$\uparrow$ / CR$\downarrow$}
			                & \textbf{SR$\uparrow$ / FR$\uparrow$ / CR$\downarrow$}
			                & \textbf{SR$\uparrow$ / FR$\uparrow$ / CR$\downarrow$} \\
			\midrule
			\multicolumn{4}{@{}l}{\emph{(a) Reported by the original authors under their own protocols; not compared with (b).}}\\
			IBVS~\citep{gupta2016ibvs}                      & 42.9 / 56.2 / 3.75 & 10.6 / 28.4 / 6.14 & 15.2 / 39.5 / 4.90 \\
			EVT~\citep{zhong2024evt}                        & 32.5 / 49.9 / 40.5 & 15.7 / 35.7 / 53.3 & 18.3 / 21.0 / 44.9 \\
			Uni-NaVid (Vicuna-7B)~\citep{zhang2024uninavid} & 25.7 / 39.5 / 41.9 & 11.3 / 27.4 / 43.5 & 8.26 / 28.6 / 43.7 \\
			TrackVLA (Vicuna-7B)~\citep{zhang2025trackvla}  & 85.1 / 78.6 / 1.65 & 57.6 / 63.2 / 5.80 & 50.2 / 63.7 / 17.1 \\
			\midrule
			\multicolumn{4}{@{}l}{\emph{(b) This work: one corrected evaluator, identical $1{,}380$ episodes; the PPO and \trace{} rows share the same five training runs.}}\\
			Qwen3-0.6B SFT base & 78.26 / 75.70 / 4.13 & 41.38 / 54.46 / 11.01 & 59.13 / 69.82 / 7.39 \\
			\quad + planner-head PPO  & 86.45{\scriptsize$\pm$1.30} / 80.88{\scriptsize$\pm$0.79} / 4.09{\scriptsize$\pm$0.40} & 43.99{\scriptsize$\pm$0.50} / 55.44{\scriptsize$\pm$0.66} / 13.78{\scriptsize$\pm$1.31} & 63.67{\scriptsize$\pm$1.04} / 72.48{\scriptsize$\pm$0.30} / 8.74{\scriptsize$\pm$0.65} \\
			\quad + \trace{} & \textbf{87.51}{\scriptsize$\pm$0.69} / \textbf{81.80}{\scriptsize$\pm$0.30} / \textbf{3.91}{\scriptsize$\pm$0.24}$^{\dagger}$ & \textbf{46.02}{\scriptsize$\pm$0.70} / \textbf{57.58}{\scriptsize$\pm$0.47} / \textbf{9.83}{\scriptsize$\pm$0.49} & \textbf{64.41}{\scriptsize$\pm$1.35} / \textbf{72.57}{\scriptsize$\pm$0.58} / \textbf{8.14}{\scriptsize$\pm$0.46}$^\star$ \\
			\midrule
			\multicolumn{4}{@{}l}{\emph{Paired effect of {\normalfont\trace{}} over planner-head PPO (intervals in Table~\ref{tab:paired_stats})}}\\
			$\Delta$ (\trace{} $-$ PPO) & $+1.06$ / $+0.92$ / $-0.17$ & $\mathbf{+2.03}$ / $\mathbf{+2.14}$ / $\mathbf{-3.96}$ & $+0.74$ / $+0.09$ / $-0.59$ \\
			\bottomrule
		\end{tabular}%
	}
	\caption{\textbf{EVT-Bench results} (SR/FR/CR, percent). Panel (a) reproduces numbers published by the original authors under their own protocols; panel (b) is measured by us under one corrected evaluator (Appendix~\ref{app:eval_protocol}). Our claims are the paired deltas of the
		last row, computed from unrounded means (they can differ from differences of the
		rounded entries by $0.01$). Bold: best within panel (b); $\pm$: std over five
		training runs; the SFT base is a single supervised model. $^{\dagger}$STT uses per-run model selection, not the gate, which fires on $0\%$ of STT steps
		(\S\ref{sec:stt_selection}); the penalty level was chosen on half of the episodes, so the unbiased STT inference is the independent-subset row of Table~\ref{tab:paired_stats}. $^\star$AT uses the AT-calibrated gate; its paired deltas did not reproduce under concurrent evaluation (Table~\ref{tab:paired_stats}).}
	\label{tab:main_compact}
\end{table*}

\begin{figure}[htbp]
	\centering
	\includegraphics[width=\linewidth]{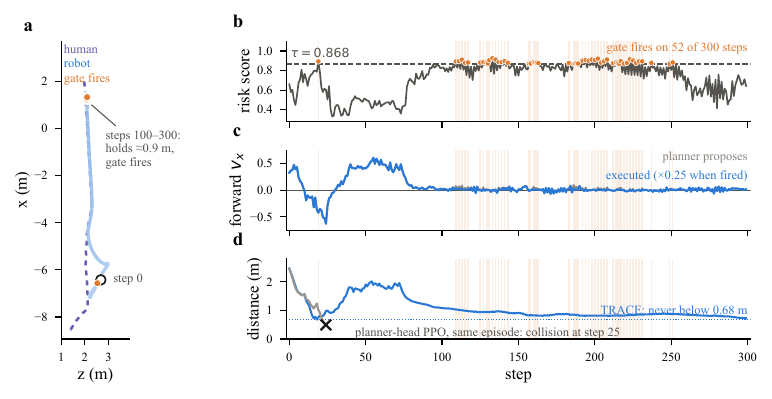}
	\caption{\textbf{One DT episode under \trace{}, on a single time axis.}
		\textbf{a}, Bird's-eye paths of the human and the robot (robot path shaded light to
		dark with time); after step $100$ the robot holds $0.7$--$1.1$\,m behind the human, and
		the fired steps cluster there. \textbf{b}, The risk score crosses $\tau{=}0.868$ on
		$52$ of $300$ steps (shaded; a close-quarters episode chosen for illustration, the
		run-wide rate is below $2\%$). \textbf{c}, Forward component of the proposed and the executed waypoint: on the $47$ fired steps moving forward the executed value is $0.25\times$ the proposed one, and on the $5$ reversing fired steps and all $248$ unfired steps the two coincide. \textbf{d}, Distance to the
		human stays above $0.68$\,m for all $300$ steps, whereas planner-head PPO on the
		same episode collides at step $25$.}
	\label{fig:qualitative_process}
\end{figure}

\begin{figure}[t]
	\centering
	\includegraphics[width=\linewidth]{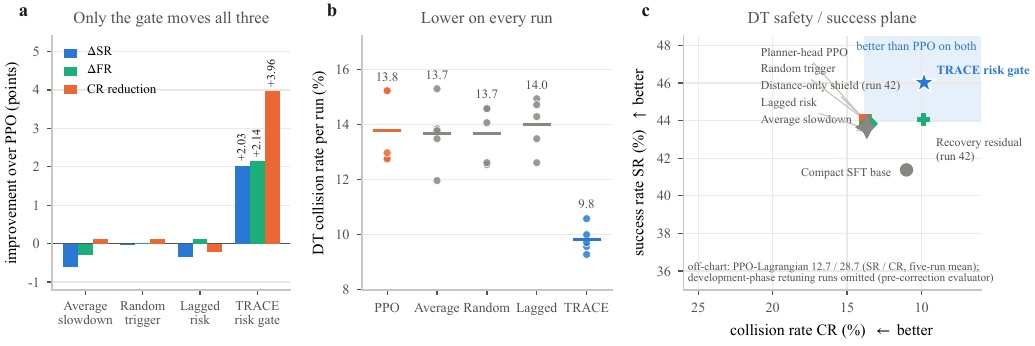}
	\caption{\textbf{\trace{} against the DT alternatives, all under the corrected
		evaluator.} \textbf{a}, Only \trace{} moves all three metrics ($+2.03$ SR,
		$+2.14$ FR, $-3.96$ CR); the matched attenuation controls stay near zero.
		\textbf{b}, Per-run CR: PPO and the three controls average $13.7$--$14.0$, while
		\trace{} is lower on every run ($9.3$--$10.6$, mean $9.83$). \textbf{c}, The DT
		success/collision plane: the controls and the diagnostic filters cluster around
		PPO, and only \trace{} reaches the high-SR, low-CR region. Single-run points are marked; all
		others are five-run means. PPO-Lagrangian lies off the chart and is quoted in the
		corner.}
	\label{fig:dt_pareto_alternatives}
\end{figure}

The collision reduction holds on each of the five runs ($9.3$--$10.6$ against $12.75$--$15.22$ for PPO; Figure~\ref{fig:dt_pareto_alternatives}b), and it is unchanged on the $281$ episodes held out from fitting the risk head ($\Delta$CR $-4.13$; Table~\ref{tab:paired_stats}). \textbf{Distracted tracking} carries the causal claim: SR $+2.03$, FR $+2.14$, and CR $-3.96$ are all significant (Table~\ref{tab:paired_stats}), and the collision reduction holds on every run (Appendix~\ref{app:identity}).
On ambiguity tracking the AT-calibrated gate stays within the benchmark's drift (Table~\ref{tab:paired_stats}).
Figure~\ref{fig:qualitative_process} shows one DT episode step by step.

\paragraph{Paired statistics.}
\label{sec:paired_stats}
All five DT runs improve on all three metrics, and each of the three gate intervals excludes zero (Table~\ref{tab:paired_stats}; the episode is the sampling unit, because the five runs share episodes). With both arms evaluated concurrently, $\Delta$SR is $+2.71$, $\Delta$FR $+2.36$, and $\Delta$CR $-3.77$, against $-0.29$ for null pairs whose two arms are identical, whose interval covers zero (Appendix~\ref{app:learned_detector}).

\begin{table}[htbp]
	\centering
		\scriptsize
	\setlength{\tabcolsep}{2.6pt}
	\begin{tabular}{llcccc}
		\toprule
		Setting & Intervention & $\Delta$SR [95\% CI] & $\Delta$FR [95\% CI] & $\Delta$CR [95\% CI] & $p$ (coll.) \\
		\midrule
		DT  & risk gate            & $\mathbf{+2.03}\,[+0.90,+3.17]$ & $\mathbf{+2.14}\,[+1.22,+3.05]$ & $\mathbf{-3.96}\,[-4.78,-3.14]$ & $<10^{-4}$ \\
		AT  & AT-calibrated gate, orig.\ eval. & $+0.74\,[-0.22,+1.70]$ & $+0.09\,[-0.50,+0.67]$ & $-0.59\,[-1.16,-0.04]$ & $0.043$ \\
		AT  & \quad concurrent re-eval. & $-0.20\,[-1.13,+0.72]$ & $-0.47\,[-1.03,+0.09]$ & $-0.06\,[-0.65,+0.51]$ & $0.90$ \\
		STT & model selection, full set & $\mathbf{+1.06}\,[+0.49,+1.61]$ & $\mathbf{+0.92}\,[+0.57,+1.29]$ & $-0.17\,[-0.52,+0.17]$ & $0.36$ \\
		STT & \quad independent subset only & $+0.78\,[0.00,+1.58]$ & $\mathbf{+0.93}\,[+0.43,+1.43]$ & $-0.20\,[-0.70,+0.29]$ & $0.49$ \\
		\bottomrule
	\end{tabular}
	\caption{\textbf{Paired statistics against planner-head PPO, with the episode as the
		sampling unit.} The five runs share the same $1{,}380$ episodes, so the $6{,}900$
		pairs are not independent: intervals are percentile bootstraps over episodes
		($10^4$ resamples, each episode carrying its five paired outcomes) and $p$ is a
		sign-flip permutation test ($2{\times}10^4$ flips) on the per-episode mean
		collision difference; the estimand is conditional on these five runs. Per-run exact McNemar $p$ values are in Table~\ref{tab:dt_taxonomy_multiseed}'s caption. Bold: interval excludes zero; intervals are uncorrected for multiple comparisons. The AT collision result of the original evaluation is left unbolded because it did not reproduce when both arms were evaluated concurrently (second AT row, five pairs; Appendix~\ref{app:calibration}). On the $281$ DT
		episodes whose rows were excluded from risk-head fitting
		(Appendix~\ref{app:calibration_detail}) the collision reduction is unchanged
		($\Delta$CR $-4.13\,[-6.26,-2.21]$, $p<10^{-4}$) while SR and FR do not fall
		($+0.07\,[-2.35,+2.56]$, $+0.99\,[-0.72,+2.79]$); the SR/FR gains are not resolved
		at that sample size. The second STT row restricts the test to the episodes outside
		each run's selection subset ($715$ distinct, $690$ per run): the following gain
		remains significant and the success gain is borderline.}
	\label{tab:paired_stats}
\end{table}

\begin{table}[t]
	\centering
	\scriptsize
	\setlength{\tabcolsep}{5pt}
	\begin{tabular}{lrrrr}
		\toprule
		Base $\rightarrow$ PPO ending & normal & lost & collision & base total \\
		\midrule
		normal    & 642 & 74  & 69  & 785 \\
		lost      & 109 & 307 & 30  & 446 \\
		collision & 33  & 15  & 101 & 149 \\
		\midrule
		PPO total & 784 & 396 & 200 & 1380 \\
		\bottomrule
	\end{tabular}
	\caption{\textbf{Episode-level ending transitions, compact base to planner-head PPO} (one matched run). $330$ episodes change ending in both directions, for a net $-50$ lost and $+51$ collision; the added collisions come mostly from episodes the base ended normally ($69$) or lost ($30$).}
	\label{tab:dt_transition}
\end{table}

\subsection{The Collision Cost of PPO}
\label{sec:ppo_cost}

Both the persistence gain and the collision cost originate in the planner head. On distracted tracking PPO raises CR from $11.01$ to $13.78$, and every one of the five runs ($12.75$--$15.22$) ends above the base.
On one matched run the endings shift without converting one-to-one (Table~\ref{tab:dt_transition}; Appendix~\ref{app:identity}, Figure~\ref{fig:main_dt_evidence}a--b).
Because recognition is frozen, the shift is attributable to the planner persisting longer on the same identity estimates. Retuning the reward does not remove this cost (\S\ref{sec:why_not_reward}), so the gate intervenes at execution time, where the failure arises.

\subsection{Target-State Timing versus Generic Attenuation}
\label{sec:why_not_reward}

\paragraph{Matched attenuation controls.}
We compare \trace{} with three controls that impose a comparable attenuation budget
without using contemporaneous target-state
risk~\citep{ames2019cbf,dalal2018safe,thananjeyan2021recoveryrl}. The
\emph{average-attenuation} control applies a constant attenuation at every step, with
its magnitude selected to match the mean attenuation induced by \trace{} over each run.
The \emph{random-trigger} control applies the attenuation factors used by \trace{} at uniformly sampled time steps, and the \emph{lagged-risk} control applies them using a risk score delayed by $37$ steps; both match the gate's intervention rate to within a few tenths of a percent, and the attenuation budgets are matched in expectation rather than on every trajectory. If generic motion attenuation or sparse intervention were sufficient, at least one control would reduce the collision rate while preserving the tracking metrics. Three findings follow from Table~\ref{tab:dt_matched_controls} and Figure~\ref{fig:dt_pareto_alternatives}. First, all three controls remain at the PPO reference on all three metrics, whereas \trace{} moves all three. The same controls remain at the PPO reference on ambiguity tracking over $15$ additional full evaluations (Appendix~\ref{app:at_controls}; per-run results in Table~\ref{tab:dt_matched_controls_per_seed}). Second, a substantially stronger uniform attenuation reduces the collision rate to $6.01$, but also decreases the success rate to $6.54$ and the following rate to $24.25$. The collision reduction therefore reflects degraded tracking rather than selective intervention. Third, a proximity-based rule operating at the gate's
intervention rate changes the collision rate by only $-0.68$, with the reported interval
including zero (Appendix~\ref{app:extra_controls}). These results indicate that the
observed improvement depends on the temporal alignment between the intervention and the
target-state risk, rather than on attenuation magnitude or intervention sparsity alone.
\begin{table}[htbp]
	\centering
	\footnotesize
	\setlength{\tabcolsep}{4pt}
	\begin{tabular}{lcccc}
		\toprule
		\textbf{DT, five runs}       & \textbf{SR $\uparrow$}  & \textbf{FR $\uparrow$}  & \textbf{CR $\downarrow$} & \textbf{Intervention rate} \\
		\midrule
		Planner-head PPO              & $43.99\pm0.50$          & $55.44\pm0.66$          & $13.78\pm1.31$           & --                    \\
		Average attenuation            & $43.39\pm0.87$          & $55.14\pm0.62$          & $13.67\pm1.19$           & $100\%$               \\
		Random-trigger attenuation       & $43.96\pm0.76$          & $55.47\pm0.43$          & $13.67\pm1.02$           & $1.6\%$               \\
		Lagged-risk attenuation (37 steps) & $43.65\pm0.99$          & $55.57\pm0.58$          & $14.00\pm0.96$           & $0.8\%$               \\
		\textbf{\trace{} risk gate}   & $\mathbf{46.02\pm0.70}$ & $\mathbf{57.58\pm0.47}$ & $\mathbf{9.83\pm0.49}$   & $1.6\%$               \\
		\bottomrule
	\end{tabular}
	\caption{\textbf{Matched attenuation controls on DT.} SR, FR, and CR are reported in percent as mean$\pm$std over five runs on the same $1{,}380$ episodes. The three controls are defined in the text; their intervention budgets are matched in expectation rather than on every trajectory.}
	\label{tab:dt_matched_controls}
\end{table}

Two further groups of alternatives appear in Figure~\ref{fig:dt_pareto_alternatives}c, with their numbers in Table~\ref{tab:neg_compact}. Modifying the
scalar objective or the optimizer does not resolve the trade-off: doubling the PPO
budget~\citep{schulman2017ppo,schulman2015gae} raises SR by $4.2$ points and CR by $2.2$ on a matched run; doubling the collision penalty~\citep{achiam2017cpo,ray2019safetygym} raises CR
by $7$ points at unchanged SR; and neither critic-free REINFORCE nor PPO-Lagrangian
reaches a usable policy. The two execution-time interventions fall short in different ways: a distance-only
learned filter~\citep{ames2019cbf,dalal2018safe,thananjeyan2021recoveryrl} does not
improve on PPO, and a supervised recovery
residual~\citep{johannink2019residual,silver2018residual} reaches a collision rate
comparable to \trace{} at lower SR and with greater variation across runs. \trace{} is
the only evaluated method that improves tracking success and collision safety together.
\paragraph{Sparse firing concentrates on high-risk steps.}
\label{sec:ablation}
The gate fires on $1.0$--$2.0\%$ of steps per run ($1.6\%$ pooled; Appendix~\ref{app:diagnostics}, Figure~\ref{fig:shield_profile}), yet touches $31\%$ of episodes, and the mean risk score on fired steps is $0.91$ against $0.36$--$0.39$ over all steps. On $68\%$ of fired steps the command is forward and is scaled to $s_f{=}0.25$; yaw is preserved, so the robot keeps facing the target during the pause and resumes at full speed once the score drops below $\tau$, which accounts for the rise in FR. Every feature group of the risk head contributes to collision prediction, with action context and identity margin contributing most (Appendix~\ref{app:diagnostics}), and the gate reduces collision-ending episodes on every one of the five runs (Appendix~\ref{app:identity}). Across thresholds $\tau\in\{0.83,\dots,0.90\}$ CR stays below PPO's, whereas the joint SR/FR/CR gain is specific to the calibrated $\tau{=}0.868$ (Appendix~\ref{app:tau_sensitivity}).

\paragraph{The gate slows near the human, not away from the scene.}
\label{sec:safety_audit}
The gate slows the robot where the human is close ($0.86$\,m median on fired steps against $1.54$\,m on unfired steps; Table~\ref{tab:safety_audit}), while per-step wall clearance matches PPO at every percentile (deltas $\leq 0.015$\,m; Appendix~\ref{app:wall_clear}).

\paragraph{What the target-state cues contribute.}
\label{sec:deployability}\label{sec:calibration_main}
The head ranks risk well across label windows (ROC-AUC $0.78$--$0.84$), with the identity group among its strongest inputs (Appendix~\ref{app:diagnostics}). Because the boxes and the association come from the benchmark's own sensors, the evidence concerns \emph{when} to intervene rather than a perception stack: with a learned COCO person detector and the probe's own association in their place, the effect is about one CR point (Appendix~\ref{app:learned_detector}).

\subsection{Robustness within EVT-Bench}
\label{sec:generalization}

Within the released environment we examine four axes (Appendix~\ref{app:generalization}). The paired DT effect holds on both scene corpora, on success as well as on collisions (HM3D~\citep{ramakrishnan2021hm3d} $\Delta$SR $+2.11$ / $\Delta$CR $-4.02$ over $3{,}360$ pairs, MP3D~\citep{chang2017matterport3d} $+1.95$ / $-3.90$ over $3{,}540$). Under three alternative collision geometries ($0.10$/$0.15$/$0.20$\,m lateral offsets against the default $0.25$\,m) the CR reduction keeps its direction ($-1.03$/$-2.15$/$-0.72$ against $-3.96$). Transferring the DT operating point to AT \emph{unchanged} lowers CR on all five runs ($8.01$ against $8.74$) at a one-point SR/FR cost that task-specific calibration recovers, and the head stack trains at three backbone scales with no backbone-dependent gap.

\section{Conclusion}
\label{sec:conclusion}

PPO on the planner head of a frozen-backbone $0.6$B VLA tracker raises success and following rates on every EVT-Bench setting, but on distracted tracking part of the added persistence turns into collisions. A target-state risk gate at execution time (\trace{}) removes this cost without removing the persistence: collisions fall by $3.96$ points while success and following rates keep rising, and matched attenuation controls without target-state timing do not reproduce this. The effect is about three times the benchmark's re-evaluation drift and reproduces under concurrent evaluation. Because the gate reads only quantities the tracker already computes and leaves the policy untouched, its operating point can be recalibrated for a new setting without retraining the planner. Target-state-aware execution therefore recovers the safety cost of policy post-training with minimal intervention.

\section*{AI Use Statement}

In this work, we used generative AI tools only to edit author-written text for
grammar, style, and readability. The authors reviewed all AI-assisted revisions and
take full responsibility for the final manuscript, including its claims, citations,
and experimental reporting.

\section*{Ethics Statement}

This work studies robots that follow people, and the deployment framing carries risks that
grow rather than shrink with the method's success. A tracker that is better at persisting
is also better at following someone who does not wish to be followed, and the risk head
depends on identity estimates that can fail asymmetrically across appearance groups; we
have not audited that asymmetry and cannot claim it is absent. All experiments here run in
simulation with synthetic humanoid avatars, so no human subjects or personal data are
involved, but the collision rates we report are simulator contact events and must not be
read as a physical-safety guarantee for a real platform. The gate offers no formal
guarantee and fails silently when it does not fire.

\section*{Reproducibility Statement}

Section~\ref{sec:setups} describes the benchmark and evaluation settings.
Appendices~\ref{app:training}, \ref{app:heads}, and~\ref{app:eval_protocol}
document model architectures, hyperparameters, training procedures, and evaluation
implementation details, including the evaluator correction. Risk-head data preparation
and partitions are detailed in Appendix~\ref{app:calibration_detail}.
Risk-feature definitions and gate thresholds appear in
Section~\ref{sec:risk_shield} and Appendices~\ref{app:feature_table}
and~\ref{app:calibration}, with sensitivity analyses and additional diagnostics in
Appendices~\ref{app:tau_sensitivity} and~\ref{app:diagnostics}.
Results measured on the diagnostic set used to calibrate the gate and on the STT
selection subsets are explicitly identified. Training and evaluation code is included
in the supplementary material and will be publicly released.

\clearpage
\bibliographystyle{iclr2027_conference}
\bibliography{example}

\clearpage
\appendix

\section{Extended Related Work}
\label{app:related}

\paragraph{Embodied tracking, following, and social navigation.}
Beyond the lineages cited in \S\ref{sec:related}, adversarial training with a learned
target~\citep{zhong2019advatplus} and the environment released with
EVT~\citep{zhong2024evt} shaped the benchmark we use. Decoupled designs have also been
built on association-based trackers~\citep{zhang2022bytetrack}, open-vocabulary
detection~\citep{liu2023groundingdino}, coarse-to-fine visual
tracking~\citep{zhang2018coarse}, transformer navigation
policies~\citep{zeng2024poliformer}, and pose-assisted
control~\citep{li2020poseassisted}. Adjacent problems share the same difficulty of
keeping a specific person: person-following with
re-identification~\citep{ye2024reid,ye2025rpfsearch}, social
navigation~\citep{mavrogiannis2023core,francis2025social,scofano2024humanthread}, and
pursuit--evasion~\citep{bajcsy2024pursuit}.

\paragraph{Post-training objectives and safe control.}
Off-policy actor--critic methods~\citep{haarnoja2018sac} and
preference- or imitation-style objectives~\citep{peters2007rwr,rafailov2023dpo,shao2024deepseekmath}
are the other common post-training routes; we report attempts with several of them in
Appendix~\ref{app:negative}. The embodied controllers post-trained this way include
navigation and manipulation policies~\citep{shah2022offlinenav,devo2021enhancing,brohan2022rt1,black2025pi05,li2024cogact,qu2025spatialvla}.
Residual policies for safety also have a supervised
variant~\citep{silver2018residual}, which we evaluate as a diagnostic in
Appendix~\ref{app:negative}.

\paragraph{Pretrained components and simulation infrastructure.}
The re-identification literature includes the association
baselines~\citep{bewley2016sort} and surveys~\citep{ye2021reidsurvey} behind the identity
quantities we consume. Frozen backbones of this kind
include~\citep{sun2023evaclip,caron2021dino,kirillov2023sam,ravi2024sam2,yang2023som,openai2024gpt4o,touvron2023llama,chiang2023vicuna,team2023gemini,liu2023visualinstruction,li2023llamavid},
and the action-head literature
includes~\citep{song2021ddim,peebles2023dit,sridhar2024nomad,ke2024dpa,liao2024diffusiondrive}.
The simulator and benchmark conventions we inherit come from
Habitat~\citep{szot2021habitat2}, vision-language
navigation~\citep{ku2020rxr,zhang2024vlnsurvey,zhou2025navgpt2,long2024instructnav},
object navigation and embodied
question answering~\citep{chaplot2020objectnav,zhang2023objectnav3d,cao2024cognav,islam2023eqamx,majumdar2024openeqa},
navigation foundation models~\citep{shah2023gnm,zhang2024navid,cheng2024navila}, and
related embodied settings~\citep{ding2024open6dor,roth2024viplanner,zhong2025dexgraspvla}.
Pedestrian-speed conventions~\citep{knoblauch1996pedestrian}, reciprocal collision
avoidance~\citep{vandenberg2011orca}, and $k$-means~\citep{arthur2007kmeans} are used in
the benchmark's motion model and in our diagnostics.

\section{Training Details}
\label{app:training}

\paragraph{Supervised tracker.}
The supervised tracker follows the TrackVLA recipe: per-frame vision tokens from a
frozen SigLIP-so400m-patch14-384 encoder concatenated with DINO-family tokens
($1{,}536$ dimensions in total) are projected by a recognition adapter into a frozen
Qwen3-0.6B language backbone, and a planner head emits $8\times3$ waypoints. The head
stack is trained by supervised fine-tuning for $2$ epochs at batch size $8$ and learning
rate $2\times10^{-5}$ on the original tracking corpus (about $1.4$M samples; the
$183$k-sample expert-rollout corpus of Appendix~\ref{app:generalization} is about $13\%$
of it), with the encoder and backbone frozen throughout.

\paragraph{Backbone and freeze policy.}
The TrackVLA-style model couples a vision encoder, a language
backbone, and a planner head producing $8\times3$ waypoints. For
post-training we freeze the vision encoder, the language backbone, and
the recognition adapter projecting visual tokens into the language space.
Only the planner head, the stochastic wrapper around it, and a learned
value head receive gradients.

\paragraph{Optimizer, rollout, and reward.}
We use AdamW (learning rate $10^{-4}$, weight decay $0.01$, grad-norm
clip $1.0$) with a constant schedule; PPO uses clip $\epsilon=0.2$,
value-loss coefficient $0.5$, entropy coefficient $0.01$, GAE
$\gamma=0.99$, $\lambda=0.95$, four PPO epochs per round, and
mini-batch size $32$. The stochastic wrapper is initialized at
$\log\sigma=-2.0$ on the full $8\times3$ waypoint head. Each round
collects $100$ on-policy episodes; the two-round main run uses $200$
episodes total. The reward of Equation~\ref{eq:reward} is a per-step
sum: follow / collision indicators read EVT-Bench's follow-distance
and contact flags, $d_t$ is the simulator-reported geodesic distance,
and the far penalty fires when $d_t > 4$\,m. Table~\ref{tab:hyper}
consolidates these settings with the evaluation budgets.

\paragraph{Hardware and wall-clock.}
Training and the full evaluations were run on a server with eight \texttt{NVIDIA A100}
($40$\,GB) GPUs. Planner-head PPO ($200$ episodes, $2$ rounds) takes ${\sim}2$\,h on one
GPU. A full $1{,}380$-episode DT evaluation is split into $30$ episode partitions; each
partition worker uses one GPU, and with the eight GPUs sharing the partitions a full
evaluation completes in ${\sim}3.5$\,h, with or without the gate. The gate runs the risk head on every step --- the score determines whether the step fires --- and on fired steps additionally rescales the executed waypoint in place. The deployed head is a $28$k-parameter multilayer perceptron (MLP) ($84 \to 128 \to 128 \to 3$ with layer normalization); a batch-1
forward pass takes ${\sim}52\,\mu$s in FP16 on an \texttt{RTX PRO 6000} datacenter GPU
and ${\sim}145\,\mu$s on one CPU core, measured over $3{,}000$ timed calls of the deployed
architecture. Both are far below the $100$\,ms period of the $10$\,Hz controller, but
neither is an end-to-end latency measurement on a robot platform, which we have not
performed.

\section{Risk Head and Recovery Residual Architecture}
\label{app:heads}

\paragraph{Inputs and heads.}
The risk head and the analysis-only recovery residual head share a
deployable target-state input (Appendix~\ref{app:feature_table}): detector-side box
geometry and confidence, top-2 identity margin, probe outputs including distance and
bearing bins, memory age and status flags, the previous command, and the planner's
predicted waypoints. The risk head is a two-layer MLP with hidden width $128$, layer
normalization, GELU activations and dropout $0.10$ ($84\to128\to128\to3$, $28$k
parameters); it emits a collision-risk score through a sigmoid, with auxiliary lost and
failure outputs under the same trunk. The residual head shares the design and emits a
three-dimensional residual on
$(v_x, v_y, \omega)$. Training uses AdamW (weight decay
$1\times10^{-4}$): the risk head with a balanced positive sampler for
up to $12$ epochs, the residual head with MSE against ground-truth deltas
(Appendix~\ref{app:residual_data}) for up to $80$ epochs, both with
early stopping on held-out validation. The residual is not part of
\trace{} and is reported only as analysis in
Appendix~\ref{app:robotframe} (Figure~\ref{fig:recovery_grounding}a).

\paragraph{Online integration.}
At deployment, the risk head consumes the same per-step target-state quantities that the diagnostic evaluation records. The gate
reads the risk score $r_t$, applies threshold $\tau$, and scales
the executed waypoint by $(s_f, s_l, s_\omega)$ when triggered; the
residual head, when run for analysis, instead adds its clipped
residual to the executed waypoint.

\section{Evaluation Protocol Details}
\label{app:eval_protocol}

\paragraph{Episode count and metrics.}
Each EVT-Bench task setting contains $1{,}380$ episodes. SR is the fraction of
episodes that satisfy the EVT-Bench success condition; FR is the
fraction of timesteps in which the agent satisfies the follow-distance
condition; CR is the fraction of episodes ending in collision.

\paragraph{Leakage-free episode resets.}
The corrected evaluator clears all episode-local state at episode boundaries, including coarse history tokens, the frame buffer, local
controller state, the last executed action, and the last predicted
trajectory. Planner failure (e.g., a tokenizer exception) is mapped to a
stop action so that the controller does not run on stale velocity
commands. We adopted this convention after observing that residual state
across episode boundaries shifts metrics by several points on DT and AT
in particular, where the policy is supposed to re-acquire the correct
target identity at the start of each new episode. All internal deltas in
the headline tables compare same-protocol numbers under this convention,
and numbers from the released (leaking) evaluator are never mixed in.

\paragraph{Reproducibility of the evaluator, and the resolution it allows.}
The evaluator is not bit-reproducible: the tracker's GPU forward pass is not
deterministic, so repeated evaluations of one policy execute slightly different
commands and diverge. Three independent evaluations of planner-head PPO run 42 under
one code version give $41.88/54.50/14.06$, $42.54/54.19/13.48$ and
$42.83/54.50/12.83$; pairwise, the same episode ends the same way in $75.3$--$77.8\%$
of cases ($90$--$91\%$ agreement on the collision flag alone), and the paired CR
differences between these identical arms are $-0.58$, $-1.23$ and $-0.65$ points, each
with an episode-bootstrap interval of about $\pm1.6$ points. Outcomes also drift across
code versions: the September re-evaluation differs from the April
$44.57/56.41/12.75$ of Table~\ref{tab:main_compact}, agreeing on $74.6\%$ of episodes.

Two consequences. First, every paired comparison must be made between evaluations under
one code version, which is how the supplementary experiments of
Appendices~\ref{app:learned_detector} and \ref{app:extra_controls} are built. Second,
with one evaluation per arm per run, differences of about one CR point or two SR points
are not resolvable. Three lines of evidence: a mis-configured gate that fires on
$0.18\%$ of steps measures $-1.07$ CR points over four runs, the same as the
learned-detector gate's $-1.09$; a null pair whose two arms are the identical
configuration, run concurrently, moves CR by $-1.38$; and two identical gate-off
evaluations of one run differ in SR by $-2.68$ with an episode-bootstrap interval of
$[-4.93,-0.43]$ that excludes zero. The episode bootstrap therefore states the
uncertainty of one pair of evaluations, not the uncertainty of the quantity those
evaluations estimate. Effects of that size in this paper are reported as unresolved. The DT risk gate's effect is three times larger, holds on every run in
both April ($-3.96$) and September ($-3.28$, Appendix~\ref{app:learned_detector}), and
concentrates where the gate fires, so it is not explained by this noise.

\paragraph{Seed convention.}
Five-seed full evaluations use seeds $\{42, 43, 44, 45, 46\}$, one per independent planner-head training run; ``seed'' and ``run'' therefore name the same unit throughout the paper. Each seed runs the same $1{,}380$ episodes; metrics across seeds are aggregated
as mean and standard deviation. All paired-seed comparisons (e.g.,
PPO~vs.\ gated PPO) use the same episode order and the same
evaluator seed on both sides. Single-seed full evaluations are tagged
with the specific seed and compared only against the same seed of the
matched baseline.

\paragraph{Logging.}
The evaluator writes per-episode JSON records with episode ID, success
flag, collision flag, follow timesteps, target-state trace, and per-step
diagnostic fields. These records are reused without modification by the ending-taxonomy, risk, and residual analyses, so that the tables below draw from the same evaluation.

\begin{table}[!htbp]
	\centering
	\scriptsize
	\setlength{\tabcolsep}{5pt}
	\renewcommand{\arraystretch}{1.0}
	\begin{tabular*}{\linewidth}{@{\extracolsep{\fill}}p{2.8cm}p{5.0cm}p{3.5cm}@{}}
		\toprule
		\textbf{Module} & \textbf{Hyperparameter}                & \textbf{Value}              \\
		\midrule
		\multirow{8}{*}{PPO planner-head}
		                & Learning rate (planner / value)        & $1\times10^{-4}$            \\
		                & Weight decay                           & $0.01$                      \\
		                & Clip $\epsilon$                        & $0.2$                       \\
		                & Discount $\gamma$                      & $0.99$                      \\
		                & GAE $\lambda$                          & $0.95$                      \\
		                & Value-loss coefficient                 & $0.5$                       \\
		                & Entropy coefficient                    & $0.01$                      \\
		                & PPO epochs / mini-batch                & $4$ / $32$                  \\
		\midrule
		\multirow{4}{*}{Stochastic wrapper}
		                & Initial $\log\sigma$                   & $-2.0$                      \\
		                & Action shape                           & $8\times3$ waypoints        \\
		                & Action parameterization                & Planar waypoints            \\
		                & Grad-norm clip                         & $1.0$                       \\
		\midrule
		\multirow{4}{*}{Reward}
		                & $w_f$ (follow)                         & $1.0$                       \\
		                & $w_c$ (collide)                        & $5.0$                       \\
		                & $w_d$ (distance shaping)               & $0.5$                       \\
		                & $w_{\mathrm{far}}$, $d_{\mathrm{far}}$ & $0.3$, $4.0$\,m             \\
		\midrule
		\multirow{3}{*}{Rollout budget}
		                & Episodes per round                     & $100$                       \\
		                & Rounds                                 & $2$                         \\
		                & Total episodes                         & $200$                       \\
		\midrule
		\multirow{4}{*}{Risk head}
		                & Hidden / depth / dropout               & $128$ / $2$ / $0.10$        \\
		                & Weight decay                           & $1\times10^{-4}$            \\
		                & Epochs (max)                           & $12$                        \\
		                & DT $(\tau, s_f, s_l, s_\omega)$        & $(0.868, 0.25, 0.60, 1.00)$ \\
		\midrule
		\multirow{2}{*}{Evaluation}
		                & Episodes per evaluation setting        & $1{,}380$                      \\
		                & Seeds (full eval)                      & $\{42,43,44,45,46\}$        \\
		\bottomrule
	\end{tabular*}
	\caption{\textbf{Centralized hyperparameters.} All values are those used for the headline numbers. Recovery-residual hyperparameters used in the bottleneck
		analysis (Appendix~\ref{app:robotframe}) are listed separately in
		Appendix~\ref{app:residual_data}.}
	\label{tab:hyper}
\end{table}

\section{Gate Activation Profile and Risk-Feature Ablation}
\label{app:diagnostics}

\begin{figure}[htbp]
	\centering
	\includegraphics[width=\linewidth]{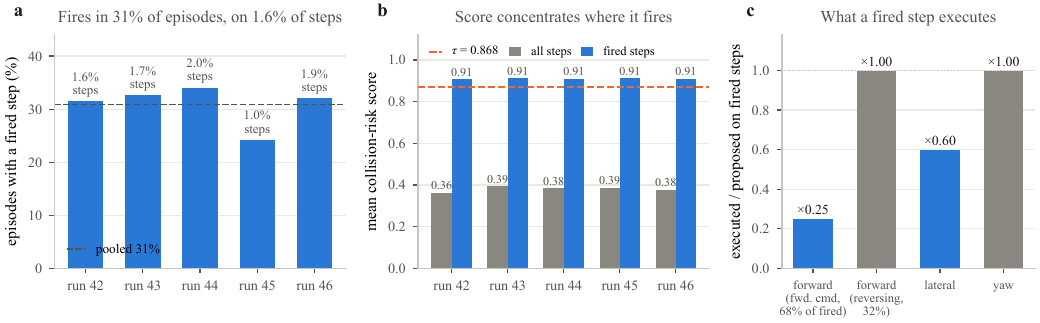}
	\caption{\textbf{Gate activation profile on DT, from the deployed step logs of the five
		runs.} \textbf{a}, Share of episodes on which the gate fires at least once (bars:
		$24$--$34\%$, $31\%$ pooled) and share of steps on which it fires (printed:
		$1.0$--$2.0\%$, $1.6\%$ pooled over $741{,}278$ steps). \textbf{b}, Mean
		collision-risk score on all steps against fired steps ($0.36$--$0.39$ against
		$0.91$; dashed line $\tau{=}0.868$). \textbf{c}, What a fired step executes: a
		forward command ($68\%$ of fired steps) is scaled to $0.25$, a reversing command is
		left at $1.0$, lateral motion is scaled to $0.60$, and yaw is unchanged.}
	\label{fig:shield_profile}
\end{figure}

\begin{figure}[htbp]
	\centering
	\includegraphics[width=\linewidth]{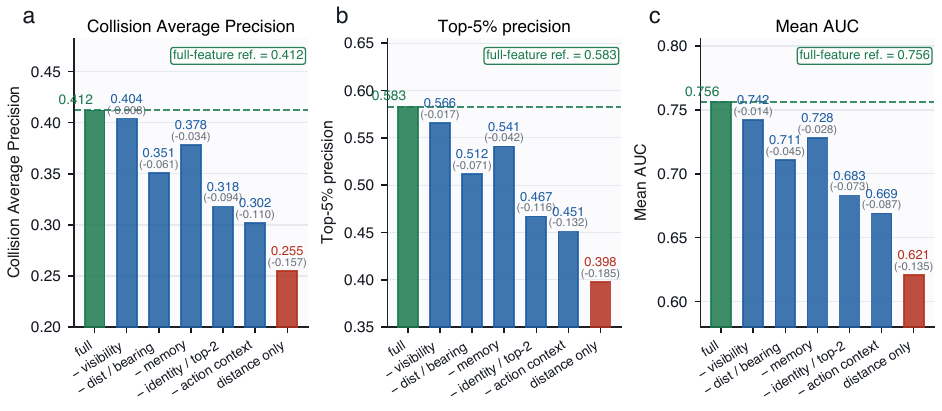}
	\caption{\textbf{Offline risk-feature ablation.} Collision-ranking
		quality---average precision, top-$5\%$ precision, and mean AUC---as inputs
		to the risk head are removed. Dropping \emph{action context} or
		\emph{identity\,/\,top-2} hurts most, and a \emph{distance-only} head is
		weakest; the ordering is consistent across all three metrics. Dashed lines
		mark the full-feature reference.}
	\label{fig:risk_feature_ablation}
\end{figure}

This appendix collects the three diagnostic figures behind \S\ref{sec:why_not_reward}. Figure~\ref{fig:dt_pareto_alternatives} places every DT alternative evaluated under the corrected evaluator on the safety/success plane; Figure~\ref{fig:shield_profile} gives the gate's activation profile from the deployed step logs of the five runs; Figure~\ref{fig:risk_feature_ablation} reports the offline feature ablation of the risk head.

\section{Matched Controls on Ambiguity Tracking}
\label{app:at_controls}

Distracted tracking carries the matched controls of \S\ref{sec:why_not_reward}; to give ambiguity tracking the same treatment we ran the same three controls on AT with the calibrated gate's risk head, threshold, and scales --- average slowdown at every step, random-trigger slowdown at approximately each run's measured gate rate (realized $0.83$--$1.23\%$ of steps against the gate's $0.76$--$1.15\%$), and lagged-risk slowdown reading the gate's score delayed by $37$ steps, which on AT crosses the threshold on $3.4$--$6.7\%$ of steps, three to six times the gate's own rate --- for $15$ further full $1{,}380$-episode evaluations (Table~\ref{tab:at_controls}). All three stay at the PPO reference, with CR within $-0.13$/$+0.07$ of $8.74$ and SR at or below PPO. In the original evaluation the gate lowered CR with non-decreasing mean SR and FR, but this margin did not reproduce when both arms were evaluated concurrently (Appendix~\ref{app:calibration}), so on AT neither the controls nor the gate shows an effect that the benchmark can resolve. Per-run values (Table~\ref{tab:at_controls_runs}) show that the original CR reduction came from runs 42 and 43 and the SR gain from runs 43 and 45, consistent with the marginal ambiguity-tracking intervals in Table~\ref{tab:paired_stats}.

\begin{table}[htbp]
	\centering
	\footnotesize
	\setlength{\tabcolsep}{4pt}
	\begin{tabular}{lcccc}
		\toprule
		\textbf{AT, five runs} & \textbf{SR $\uparrow$} & \textbf{FR $\uparrow$} & \textbf{CR $\downarrow$} & \textbf{Intervention rate} \\
		\midrule
		Planner-head PPO              & $63.67\pm1.04$ & $72.48\pm0.30$ & $8.74\pm0.65$ & -- \\
		Average attenuation              & $63.46\pm0.88$ & $71.99\pm0.70$ & $8.81\pm0.50$ & $100\%$ \\
		Random-trigger attenuation & $63.62\pm1.84$ & $72.27\pm1.04$ & $8.61\pm0.95$ & $0.83$--$1.23\%$ \\
		Lagged-risk attenuation (37 steps) & $62.91\pm1.77$ & $71.78\pm1.10$ & $8.62\pm1.05$ & $3.4$--$6.7\%$ \\
		\textbf{\trace{} calibrated gate} & $\mathbf{64.41\pm1.35}$ & $\mathbf{72.57\pm0.58}$ & $\mathbf{8.14\pm0.46}$ & $0.76$--$1.15\%$ \\
		\bottomrule
	\end{tabular}
	\caption{\textbf{Matched slowdown controls on AT.} SR/FR/CR in percent, mean$\pm$std
		over the five training runs on the identical $1{,}380$ episodes; controls reuse the calibrated gate's risk head, threshold, and scales and differ in the trigger
		source. Budgets are comparable, not identical (intervention-rate column).}
	\label{tab:at_controls}
\end{table}

\begin{table}[htbp]
	\centering
	\scriptsize
	\setlength{\tabcolsep}{2.5pt}
	\begin{tabular}{lcccccc}
		\toprule
		\textbf{Run} & \textbf{PPO} & \textbf{Average} & \textbf{Random} & \textbf{Lagged} & \textbf{Calibrated gate} & \textbf{Gate rate} \\
		\midrule
		42 & $62.54/72.52/9.06$ & $62.90/71.04/8.55$ & $62.03/71.35/8.26$ & $61.30/70.83/9.20$ & $63.33/71.90/8.04$ & $0.98\%$ \\
		43 & $63.04/72.12/9.49$ & $64.71/72.55/8.84$ & $64.93/73.23/9.13$ & $63.62/72.38/8.70$ & $66.30/73.22/7.83$ & $0.92\%$ \\
		44 & $65.14/72.66/8.99$ & $63.55/71.96/9.64$ & $63.19/72.33/9.86$ & $63.33/71.62/10.00$ & $63.91/72.63/8.91$ & $1.15\%$ \\
		45 & $63.33/72.83/7.90$ & $63.77/72.77/8.33$ & $66.09/73.35/7.32$ & $65.29/73.35/7.39$ & $65.29/73.04/7.75$ & $0.76\%$ \\
		46 & $64.28/72.24/8.26$ & $62.39/71.65/8.70$ & $61.88/71.08/8.48$ & $61.01/70.73/7.83$ & $63.19/72.05/8.19$ & $1.08\%$ \\
		\bottomrule
	\end{tabular}
	\caption{\textbf{AT controls per run} (SR/FR/CR in percent, $1{,}380$ episodes each).
		The five control evaluations per row share one training run; the gate's gain is not uniform across runs. The last column is the gate's measured step-trigger rate on that run; the random control is matched to it approximately (realized $1.00/1.23/0.90/0.83/1.08\%$), and the lagged control fires on $5.4/5.1/6.7/3.4/5.6\%$ of steps.}
	\label{tab:at_controls_runs}
\end{table}

\section{Perception: Replacing the Benchmark's Detection Stream}
\label{app:learned_detector}

The headline gate's candidate boxes and target association come from EVT-Bench's
panoptic humanoid sensors (Appendix~\ref{app:feature_table}). To measure what the gate
does with realistic perception we replaced that stream end-to-end and re-ran the DT
protocol.

\textbf{Perception.} Candidates are the \emph{person} detections of an RT-DETR
(ResNet-50, COCO) detector run on the egocentric RGB frame (score $\ge 0.30$, at most
$20$ boxes ordered by area); the target is the candidate selected by the SigLIP
grounding probe from the instruction text, exactly as in the headline system, and the
sixteen detector features are recomputed from that selection: the selected box, its
probe confidence in place of the benchmark-box overlap, the probe's top-2 margin, and
the statistics of the probe's candidate confidences. The visible-human count and the
image-plane flags follow the same boxes. The detector finds a single person more often
than the panoptic enumeration, so the probe, which needs two candidates, is unavailable
on $44.6\%$ of steps against $32.2\%$ before; we kept that requirement unchanged.

\textbf{Head and threshold.} Features were logged from the five planner-head PPO runs
with the gate disabled ($143{,}277$ rows, stride $5$, at most $80$ rows per episode, the
same $80/20$ episode-hash split as Appendix~\ref{app:calibration_detail}) and the same
$84$-input head was retrained; its validation ranking is close to the original head's
on the same kind of rows (collision ROC-AUC $0.690$ against $0.713$, AP $0.112$ against
$0.146$). The threshold was set before any gated evaluation by matching the original
head's validation trigger rate ($2.0\%$ of validation rows at $\tau{=}0.868$), giving
$\tau{=}0.798$; two neighbours ($0.785$ at $2.5\%$, $0.810$ at $1.5\%$) were evaluated on run 42 and the a priori rule of Appendix~\ref{app:calibration} (minimum CR on the held-out-hash episodes subject to an SR drop of at most $0.5$ points and a CR drop of at least $1.5$) selected $\tau{=}0.798$: on those $281$ episodes the three thresholds give CR $10.32$ / $13.52$ / $11.74$ against $12.10$ for PPO, with SR $-0.36$ / $-0.36$ / $0.00$ (full-set triplets $43.12/54.98/11.67$, $42.97/54.07/13.99$, $43.84/55.70/12.46$); the scales are the DT scales.

\textbf{Same-code pairing and replication of the headline effect.} Outcomes drift
across code versions (Appendix~\ref{app:eval_protocol}), so every comparison here is
paired against the same policy evaluated under the same code with the gate disabled (the
feature-logging runs), never against Table~\ref{tab:main_compact}. Under that pairing the
\emph{original} gate, with the benchmark's panoptic perception and the deployed head,
reproduces the headline effect on all five runs: $\Delta$SR $+2.19\,[+1.10,+3.26]$,
$\Delta$FR $+1.88\,[+1.03,+2.74]$, $\Delta$CR $-3.28\,[-4.13,-2.42]$ ($p<10^{-4}$; per
run $-3.91$, $-3.26$, $-3.48$, $-2.83$, $-2.90$), firing on $1.5\%$ of steps and $30\%$
of episodes. One caveat of provenance: a May change to a feature helper altered how the
probe's normalized box is expanded into columns, which shifts the deployed head's input
distribution; the replication above restores the April computation, and without that
restoration the same head fires on only $0.18\%$ of steps. That mis-configured variant
serves as a null control below.

\begin{table}[htbp]
	\centering
	\footnotesize
	\setlength{\tabcolsep}{4pt}
	\begin{tabular}{lcccc}
		\toprule
		\textbf{Run} & \textbf{PPO, same code} & \textbf{Learned-detector gate} & \textbf{$\Delta$SR / $\Delta$FR / $\Delta$CR} & \textbf{Trigger (steps / ep.)} \\
		\midrule
		42 & $41.88/54.55/14.06$ & $43.12/54.98/11.67$ & $+1.23$ / $+0.43$ / $-2.39$ & $2.06\%$ / $44.4\%$ \\
		43 & $44.20/54.96/13.99$ & $44.49/55.63/13.77$ & $+0.29$ / $+0.67$ / $-0.22$ & $2.48\%$ / $44.1\%$ \\
		44 & $44.06/56.55/13.70$ & $45.00/57.56/13.12$ & $+0.94$ / $+1.01$ / $-0.58$ & $2.06\%$ / $44.2\%$ \\
		45 & $43.99/55.25/12.68$ & $45.07/56.40/11.38$ & $+1.09$ / $+1.16$ / $-1.30$ & $1.88\%$ / $44.5\%$ \\
		46 & $43.26/54.87/13.55$ & $43.04/56.59/12.61$ & $-0.22$ / $+1.72$ / $-0.94$ & $2.21\%$ / $45.9\%$ \\
		\midrule
		pooled & $43.48/55.24/13.59$ & $44.14/56.23/12.51$ & $+0.67$ / $+1.00$ / $-1.09$ & $2.1\%$ / $44.6\%$ \\
		\bottomrule
	\end{tabular}
	\caption{\textbf{Gate with a learned detector, per run} (SR/FR/CR in percent, full
		$1{,}380$-episode DT set, corrected evaluator, current code). The reference is
		planner-head PPO evaluated under the same code with the gate disabled.}
	\label{tab:learned_detector}
\end{table}

\textbf{First measurement, and why it does not survive.} Table~\ref{tab:learned_detector}
gives the five runs as first measured, each treatment evaluated after its reference.
Pooled and paired at the episode level the gate changes SR by $+0.67\,[-0.35,+1.78]$, FR
by $+1.00\,[+0.23,+1.77]$ and CR by $-1.09\,[-1.80,-0.36]$ (sign-flip $p{=}0.004$),
firing on $2.1\%$ of steps and $45\%$ of episodes against $1.5\%$ and $30\%$ for the
original gate. Read alone this is a reduced but surviving effect. Two null measurements
say otherwise. Repeated evaluations of one \emph{unchanged} policy differ by $-0.58$,
$-1.23$ and $-0.65$ CR points (Appendix~\ref{app:eval_protocol}), and the mis-configured
gate above --- $0.18\%$ of steps, $9\%$ of episodes --- measures $\Delta$CR $-1.07$ over
four runs, indistinguishable from the learned-detector gate's $-1.09$, with its apparent
gain sitting on the episodes it never touched. Because every treatment was evaluated
after its reference, an evaluator drift of this size is not separable from a treatment
effect.

\textbf{Time-matched re-test.} We therefore re-ran the comparison with both arms
\emph{concurrent}: the $30$ splits of the reference and the $30$ splits of the gated arm
are interleaved over the same fourteen worker slots, so both experience the same machine
state and the same wall-clock window. We added null pairs in which the two arms are the
identical gate-off configuration, which calibrates the protocol on the same axis.
Table~\ref{tab:paired_retest} gives all seven pairs. Pooled over the five runs the gated
arm moves CR by $-0.22\,[-0.96,+0.52]$ ($p{=}0.59$), SR by $-0.54\,[-1.62,+0.51]$ and FR
by $-0.27\,[-1.07,+0.52]$; the two null pairs, whose arms differ in nothing, move CR by
$-0.29\,[-1.52,+0.98]$ and SR by $+0.11\,[-1.63,+1.81]$. Treatment and null are
statistically indistinguishable, and the first measurement's $-1.09$ is accounted for by
the ordering of the two arms rather than by the gate. The floor is visible directly in
the per-pair numbers: identical configurations move CR by $-1.38$ and $+0.80$ on the two
null pairs, and three evaluations of one unchanged run-42 configuration give SR $45.29$,
$42.61$, $42.54$ and CR $12.68$, $14.06$, $12.68$.

\begin{table}[htbp]
	\centering
	\scriptsize
	\setlength{\tabcolsep}{3.5pt}
	\begin{tabular}{llccc}
		\toprule
		\textbf{Pair (concurrent arms)} & \textbf{Run} & \textbf{Arm A} & \textbf{Arm B} & \textbf{$\Delta$SR / $\Delta$FR / $\Delta$CR} \\
		\midrule
				gate-off vs learned gate & 42 & $45.29/56.40/12.68$ & $42.54/54.82/13.41$ & $-2.75$ / $-1.58$ / $+0.72$ \\
		gate-off vs learned gate & 43 & $43.91/54.84/15.43$ & $42.32/54.57/14.93$ & $-1.59$ / $-0.27$ / $-0.51$ \\
		gate-off vs learned gate & 44 & $44.49/55.22/14.28$ & $43.12/55.16/13.62$ & $-1.38$ / $-0.06$ / $-0.65$ \\
		gate-off vs learned gate & 45 & $42.46/55.32/13.12$ & $44.28/55.47/12.97$ & $+1.81$ / $+0.14$ / $-0.14$ \\
		gate-off vs learned gate & 46 & $44.06/55.59/13.62$ & $45.29/55.99/13.12$ & $+1.23$ / $+0.40$ / $-0.51$ \\
		\emph{pooled, five runs} & & & & $\mathbf{-0.54}$ / $\mathbf{-0.27}$ / $\mathbf{-0.22}$ \\
		\midrule
		gate-off vs gate-off (null) & 42 & $42.61/55.04/14.06$ & $42.54/54.63/12.68$ & $-0.07$ / $-0.41$ / $-1.38$ \\
		gate-off vs gate-off (null) & 43 & $43.26/54.09/14.49$ & $43.55/55.40/15.29$ & $+0.29$ / $+1.31$ / $+0.80$ \\
		\emph{pooled, two null pairs} & & & & $\mathbf{+0.11}$ / $\mathbf{+0.45}$ / $\mathbf{-0.29}$ \\
		\bottomrule
	\end{tabular}
		\caption{\textbf{Time-matched paired re-test} (SR/FR/CR, percent, full $1{,}380$-episode
		DT set, seven paired evaluations of $2{,}760$ episodes each). Both arms of a row run
		concurrently over the same worker slots. The null rows' two arms are the same
		configuration, so their spread is the protocol's resolution: the gated arm's pooled
		deltas sit inside it.}
	\label{tab:paired_retest}
\end{table}

\textbf{The same protocol on the original gate.} Applied to the gate of
\S\ref{sec:risk_shield} --- the benchmark's own detection, the deployed head,
$\tau{=}0.868$ --- the concurrent protocol reproduces the headline effect on all five
runs (Table~\ref{tab:dt_retest}): $\Delta$SR $+2.71\,[+1.51,+3.88]$, $\Delta$FR
$+2.36\,[+1.40,+3.29]$, $\Delta$CR $-3.77\,[-4.67,-2.90]$, sign-flip $p<10^{-5}$, against
$-0.29\,[-1.52,+0.98]$ for the null pairs. Five of five runs improve on all three
metrics, and the effect is $2.7$ times the null spread, so it is reported as a result whereas the learned-detector variant is not.

\begin{table}[htbp]
	\centering
	\scriptsize
	\setlength{\tabcolsep}{3.5pt}
	\begin{tabular}{lccc}
		\toprule
		\textbf{Run} & \textbf{Gate off} & \textbf{Risk gate ($\tau{=}0.868$)} & \textbf{$\Delta$SR / $\Delta$FR / $\Delta$CR} \\
		\midrule
		42 & $43.41/55.33/13.33$ & $44.42/56.02/9.28$ & $+1.01$ / $+0.69$ / $-4.06$ \\
		43 & $44.06/54.67/14.49$ & $45.72/56.61/11.30$ & $+1.67$ / $+1.94$ / $-3.19$ \\
		44 & $43.33/55.02/14.93$ & $47.75/58.24/10.29$ & $+4.42$ / $+3.22$ / $-4.64$ \\
		45 & $43.84/55.34/12.83$ & $46.59/57.77/8.70$ & $+2.75$ / $+2.43$ / $-4.13$ \\
		46 & $43.04/54.61/13.04$ & $46.74/58.14/10.22$ & $+3.70$ / $+3.53$ / $-2.83$ \\
		\midrule
		\emph{pooled} & $43.54/55.00/13.72$ & $46.25/57.36/9.96$ & $\mathbf{+2.71}$ / $\mathbf{+2.36}$ / $\mathbf{-3.77}$ \\
		\bottomrule
	\end{tabular}
	\caption{\textbf{The DT risk gate under the concurrent protocol} (SR/FR/CR, percent,
		full $1{,}380$-episode set, five paired evaluations of $2{,}760$ episodes each).
		Both arms of a row are evaluated at the same time on the same worker slots, the
		design that removes the drift visible in Table~\ref{tab:paired_retest}'s null pairs.}
	\label{tab:dt_retest}
\end{table}

What this establishes is a boundary on the benchmark rather than a property of the
method: with EVT-Bench's own detection replaced, the question moves to a scale the
benchmark does not measure, since one evaluation per arm resolves neither one CR point
nor two SR points. A perception-level claim therefore needs either an evaluator with
repeated measurements per arm or an effect several times larger than the one at issue.
The gate's own effect is that large --- three to four times the floor, concentrated on
the episodes it fires in ($-6.2$ CR points there against $-2.0$ elsewhere), and replicated across code versions, so it is reported as a result and the learned-detector variant as the next experiment.

\section{Strong Slowdown and a Proximity Rule}
\label{app:extra_controls}

The matched controls of \S\ref{sec:why_not_reward} test whether the gate's small
attenuation budget is sufficient without its timing. Two further controls, run under the
current code and paired against the same-code planner-head PPO reference of
Appendix~\ref{app:learned_detector}, test the converse.

\textbf{Strong uniform slowdown.} Applying the gate's own scales $(s_f, s_l, s_\omega) = (0.25, 0.60, 1.00)$ at every step, on all five runs, lowers CR from $13.59$ to $6.01$ but drives SR from $43.48$ to $6.54$ and FR from $55.24$ to $24.25$ (paired $\Delta$SR $-36.9\,[-39.1,-34.7]$, $\Delta$FR $-31.0\,[-33.0,-29.0]$, $\Delta$CR $-7.6\,[-8.9,-6.4]$; per run $6.52/24.09/5.87$, $6.67/24.87/5.80$, $6.30/23.72/5.80$, $6.59/24.06/6.45$, $6.59/24.53/6.16$). Uniform slowing therefore lowers the collision rate without tracking, whereas the timing of the gate's intervention lowers it while the following rates are preserved.

\textbf{Rule-based proximity trigger.} The converse control replaces the learned head by
a hand-written rule with the same scales: fire when the learned probe places the target
in its nearest distance bin with distance confidence at least $0.588$, a threshold chosen
on the run-42 logs to match the gate's $1.6\%$ step rate (realized $1.5$--$1.7\%$ of steps and $35$--$36\%$ of episodes). Over the five runs the rule changes SR by
$+0.17\,[-0.86,+1.22]$, FR by $+0.27\,[-0.48,+1.04]$, and CR by $-0.68\,[-1.46,+0.12]$
(sign-flip $p{=}0.10$), with CR falling on four runs and rising on one (per run
$-1.38$, $+0.36$, $-0.29$, $-1.09$, $-1.01$). Proximity alone, at the gate's budget, does not produce a resolved safety effect. It is also at the noise floor of Appendix~\ref{app:eval_protocol}, so this control bounds the effect of such a rule rather than showing it to be zero; the comparison it supports is with the original gate's $-3.28$ under the same pairing, not with the learned-detector configuration, which is itself unresolved.

\section{Generalization Evidence in Detail}
\label{app:generalization}

We probe four axes on which the effect could plausibly fail inside the released environment. None substitutes for a second benchmark or a real robot.

\textbf{Scene distribution.} EVT-Bench episodes span two independent scene corpora.
Recomputing the paired DT comparison separately on each shows the safety--success pair on
both: HM3D $\Delta$SR $+2.11$ / $\Delta$CR $-4.02$ ($3{,}360$ pairs), MP3D $+1.95$ /
$-3.90$ ($3{,}540$ pairs). The effect is not carried by one scene distribution.

\textbf{Collision definition.} Because CR depends on the simulator's contact geometry, we
re-evaluated PPO against the gate under three alternative navmesh footprints
($0.10$/$0.15$/$0.20$\,m lateral offsets, against the default $0.25$\,m; $16$ full evaluations over four, three, and one training runs respectively). The CR reduction keeps its direction ($\Delta$CR $-1.03$/$-2.15$/$-0.72$ against $-3.96$; $\Delta$SR $+0.60$/$+0.48$/$-0.14$ against $+2.03$); its magnitude does not vary monotonically with the footprint and the $0.20$\,m point is a single run, so the direction holds under every tested collision definition.

\textbf{Operating-point transfer.} Applying the DT-calibrated gate to AT
\emph{unchanged} --- same risk head, same $\tau$, same scales --- still reduces CR on all
five runs ($8.01$ against PPO's $8.74$), at a one-point SR/FR cost that
task-specific calibration recovers. What transfers is the interface; calibration is a knob
that trades following performance back.

\textbf{Backbone.} The head-stack trains and evaluates end-to-end on three backbone scales
(Qwen2.5-0.5B, Qwen3-0.6B, Qwen2.5-1.5B) with no code changes (Table~\ref{tab:backbones}).
To separate backbone from data we trained all three on one shared self-distilled corpus
($446$k samples, one epoch): they land within $1.3$ DT-SR points of each other
($7.54$ / $7.03$ / $6.30$) and far below the main model trained on the original corpus
($41.38$), which is consistent with the corpus, not the backbone, explaining the gap,
although it does not exclude backbone effects at full corpus scale. Planner-head PPO runs
end-to-end on the new backbone but does not rescue a weak base ($3.70$ DT SR). To probe
corpus scale directly we re-collected an expert-rollout corpus of $3{,}121$ episodes
($183$k samples, about $13\%$ of the original corpus) and retrained Qwen2.5-0.5B under the
original recipe ($2$ epochs, batch $8$, learning rate $2{\times}10^{-5}$): STT reaches
$55.36 / 64.35 / 4.28$, a collision rate already at the main model's $4.13$, and DT
$15.58 / 38.50 / 13.41$, twice the distilled-corpus result. These are intermediate numbers
on a partial corpus: they support the reading that corpus scale is the remaining factor,
and they are evidence of portability, not a cross-backbone performance claim, which
requires the original full-scale corpus.

\begin{table}[htbp]
	\centering
	\footnotesize
	\setlength{\tabcolsep}{3pt}
	\begin{tabular}{llcc}
		\toprule
		\textbf{Backbone} & \textbf{SFT corpus} & \textbf{DT SR / FR / CR} & \textbf{STT SR / FR / CR} \\
		\midrule
		Qwen3-0.6B (main model) & original & $41.38 / 54.46 / 11.01$ & $78.26 / 75.70 / 4.13$ \\
		Qwen2.5-0.5B & distilled, $446$k & $7.54 / 26.14 / 8.19$ & -- \\
		Qwen3-0.6B (control) & distilled, $446$k & $7.03 / 25.58 / 7.25$ & -- \\
		Qwen2.5-1.5B & distilled, $446$k & $6.30 / 23.77 / 8.91$ & -- \\
		Qwen2.5-0.5B + PPO & distilled, $446$k & $3.70 / 18.86 / 5.14$ & $6.38 / 21.75 / 4.86$ \\
		Qwen2.5-0.5B & expert rollouts, $183$k & $15.58 / 38.50 / 13.41$ & $55.36 / 64.35 / 4.28$ \\
		\bottomrule
	\end{tabular}
	\caption{\textbf{Backbone portability.} All rows are single training runs (one SFT model each, plus one PPO run) evaluated on the full $1{,}380$-episode settings under the corrected evaluator; they are not
		five-run means. The three distilled-corpus rows share one corpus and recipe.}
	\label{tab:backbones}
\end{table}

\section{Risk-Head Reliability and Recalibration}
\label{app:calibration_detail}

\begin{table}[htbp]
	\centering
	\footnotesize
	\setlength{\tabcolsep}{5pt}
	\begin{tabular}{lccccc}
		\toprule
		Label window $K$ & Positive rate (count) & ROC-AUC & AP & ECE (raw) & Brier (raw) \\
		\midrule
		$10$ & $0.89\%$ ($6{,}583$) & $0.842$ & $0.080$ & $0.371$ & $0.201$ \\
		$20$ & $1.60\%$ ($11{,}897$) & $0.813$ & $0.103$ & $0.364$ & $0.200$ \\
		$40$ & $2.50\%$ ($18{,}543$) & $0.781$ & $0.120$ & $0.355$ & $0.198$ \\
		\bottomrule
	\end{tabular}
	\caption{\textbf{Ranking quality and probability calibration of the deployed risk
		head}, over $741{,}278$ scored steps from the five deployed runs. A positive label is a
		step inside the $K$-step window preceding a collision termination. Ranking is
		strong at every window; the raw sigmoid is badly calibrated as a probability at
		every window. We therefore report the output as a risk score.}
	\label{tab:calibration}
\end{table}

Fitting an isotonic regression on one run and applying it to the four held-out runs gives
ECE $0.0016/0.0009/0.0013/0.0005$ and Brier ${\approx}0.015$: a roughly two-hundred-fold
reduction in ECE and a thirteen-fold reduction in Brier. The deployed system keeps the raw
score and the raw threshold $\tau = 0.868$ (which the map sends to a calibrated risk of
$0.112$), so no trigger decision reported in this paper depends on the map; we give it for
interpretability only.

\paragraph{Training data, partitions, and the scope of the deployed evaluation.}
The deployed head is trained on $142{,}976$ step rows logged at stride $5$ (at most $80$
rows per episode) from five full DT evaluation runs of earlier policies --- the compact
base, planner-head PPO (run 42), two distance-only filters, and an image-plane filter ---
all on the same $1{,}380$ DT episodes used for every evaluation in this paper. Training
labels are episode-terminal: every logged step of an episode that ends in collision
(lost, failure) is a positive for the collision (lost, failure) output; the $K$-step
windows of Table~\ref{tab:calibration} are used only to measure the deployed score
afterwards. Rows are split by a stable hash of the episode identifier into $80\%$
training and $20\%$ validation ($30{,}207$ validation rows, collision base rate
$4.9\%$); the threshold and scales are chosen on the validation rows' predictions by an
intervention-rate sweep (Appendix~\ref{app:calibration}) and frozen before any full
five-run evaluation. Table~\ref{tab:risk_partitions} summarizes the sets.

\begin{table}[htbp]
	\centering
	\footnotesize
	\setlength{\tabcolsep}{5pt}
	\begin{tabular}{llrrl}
		\toprule
		Set & Source & Episodes & Rows / steps & Used for \\
		\midrule
		head training   & 5 DT runs, $80\%$ of episode ids by hash & ${\approx}1{,}104$ & $112{,}769$ rows & fitting $f_\phi$ \\
		head validation & same runs, remaining $20\%$ & ${\approx}276$ & $30{,}207$ rows & $\tau$, $(s_f,s_l,s_\omega)$ \\
		deployed evaluation & 5 PPO runs $\times$ all episodes & $1{,}380$ & $741{,}278$ steps & Tables~\ref{tab:main_compact}--\ref{tab:paired_stats} \\
		\bottomrule
	\end{tabular}
	\caption{\textbf{Risk-head data partitions.} The deployed-evaluation episode set
		contains both partitions above, so the five-run results measure the gate on episodes whose scenes and identities the head saw during training under other policies' rollouts (and, for run 42, under a diagnostic pass of the same PPO policy).}
	\label{tab:risk_partitions}
\end{table}

Consequently the head's generalization to unseen episodes is supported only by the
validation partition, and the five-run deployed evaluation is a within-benchmark
measurement: it establishes what the gate does on this benchmark, not how the head would
transfer to scenes or identities it has never scored. The scene-corpus split of
Appendix~\ref{app:generalization} does not remove this overlap, because both corpora
contribute training episodes. We state this as a limit of the evidence rather than as a detected leak; the paired comparison against PPO on the same episodes remains valid as a measurement of behavior on the benchmark, and on the $281$ episodes of the validation partition, whose rows never entered fitting, the collision reduction ($\Delta$CR $-4.13\,[-6.26,-2.21]$) matches the full-set effect while the success and following gains ($+0.07$, $+0.99$) are not resolved at that sample size.

\section{Deployed Risk-Feature Definitions}
\label{app:feature_table}

The deployed head consumes $84$ scalars, all available before the step's action is
executed. They fall into eight groups. \emph{Nine action components} (the previous
command, the planner's proposed command, and the previously executed command) and
\emph{three base-velocity components}, from the policy. \emph{Sixteen detector
quantities}: candidate boxes are the image-plane boxes of humanoid instances in
EVT-Bench's panoptic segmentation, the selected box is the candidate with the highest
overlap with the benchmark's target box, its score is that overlap, the top-2 identity
margin is the gap to the runner-up, and four statistics summarize the candidate scores
(count, maximum, mean, top-2 margin); detection and target association are therefore
simulator-perfect in this study. \emph{Six image-plane quantities} (box size, centering
error, and two image-plane gate flags) computed on the selected box. \emph{Fifteen
planner outputs} (the twelve-dimensional trajectory-head summary and the first
predicted waypoint). \emph{Thirty-three probe outputs} from a learned SigLIP grounding
probe run on RGB crops of the candidates (angle and distance bins with their
confidences, selection confidence and top-2 margin, memory age, recovery and status
flags, and selected-box geometry in pixel and normalized coordinates). A
\emph{visible-human count} from the same panoptic enumeration, and the \emph{step
index}. The logged feature set also contains quantities that become available only after the
step is executed; these are not inputs of the deployed head.

No feature reads simulator pose, geodesic distance, or the benchmark following bands.
These privileged quantities are excluded from the head's inputs by construction. The detector group is a different case: it is deployable in form (boxes and scores), but
its quality is that of the simulator, because the benchmark supplies detection and
identity from its own sensors rather than from a learned detector, and Figure~\ref{fig:risk_feature_ablation}
shows the identity group is among the head's strongest inputs. Replacing it with a
learned detector and re-identifier is the step from this study to a deployable system.

\section{Risk Gate Calibration Protocol}
\label{app:calibration}

\paragraph{Diagnostic set.}
Calibration proceeds in two steps on data logged before any five-run evaluation.
First, an offline intervention-rate sweep on the $30{,}207$ validation rows of the
risk-head dataset (Appendix~\ref{app:calibration_detail}) ranks rows by score and reads
collision precision, lift, and episode capture at the top $1$--$20\%$ of rows, which
shortlists thresholds whose intervention rate is a few percent of steps. Second, the
shortlisted $(\tau, s_f, s_l, s_\omega)$ settings are checked closed-loop on a
diagnostic set built from a single full DT run under the corrected protocol, with
$20\%$ of its episodes as a tuning subset and the remaining $80\%$ as a verification sample for gate decisions before any full five-run evaluation. The
diagnostic episodes belong to the same $1{,}380$-episode benchmark as the five-run
evaluations; what is held out is the evaluated policies' rollouts, not the scenes
(the overlap is stated in Appendix~\ref{app:calibration_detail}).

\paragraph{Threshold sweep.}
The collision-risk threshold $\tau$ is swept over a coarse grid from $0.50$ to $0.90$ and then refined near the knee of the trigger-rate curve, where $0.868$ was selected; the scale triple $(s_f, s_l, s_\omega)$ is swept on a small Cartesian grid that
prefers slow-down on the forward axis and preserves angular freedom.
The selected DT setting
$(\tau, s_f, s_l, s_\omega) = (0.868, 0.25, 0.60, 1.00)$ is the one that
minimizes CR on the tuning subset subject to a non-decrease constraint
on SR.

\paragraph{AT calibration.}
The DT setting is too aggressive for AT: it lowers FR more than it lowers CR. For AT we therefore use a separate risk head and freeze two AT-specific operating points before the
full five-run evaluation. The AT head takes an $80$-dimensional input built from the
robot-frame probe features (Appendix~\ref{app:robotframe}) and is trained on rollouts
over AT episode partitions $8$--$29$, with a collision label at horizon $H{=}20$ and
lost-target and failure labels at $H{=}40$. Its decision variable is a benefit margin,
the collision score minus $0.35$ times the lost-target score, and the gate fires only
when, in addition, the selection confidence is at least $0.70$, the top-2 identity
margin is at least $0.12$, and the commanded forward speed is at least $0.01$. The
training partitions and the evaluation set are drawn from the same released AT
episodes, so the AT head is not episode-disjoint from its evaluation. The calibrated
gate uses $(\tau, s_f, s_l, s_\omega) = (0.586, 0.65, 0.85, 1.00)$, selected on the AT
tuning subset for all-metric improvement under a non-worsening CR criterion; the margin
gate uses $(0.904, 0.70, 0.90, 1.00)$, selected for higher FR under a matching CR
constraint. Both attenuate more gently than the DT gate, and the calibrated gate fires at a much lower score, so it triggers on $0.76$--$1.15\%$ of AT steps.
Table~\ref{tab:at_operating_points} reports both full five-run operating points.

\paragraph{Concurrent re-evaluation of the AT gate.}
A single evaluation per arm cannot resolve differences near one CR point
(Appendix~\ref{app:eval_protocol}). We therefore re-evaluated the calibrated gate
against its PPO reference with both arms running concurrently on the same worker
slots, five pairs with the full $1{,}380$ AT episodes each. Pooled over episodes,
$\Delta$SR is $-0.20\,[-1.13,+0.72]$, $\Delta$FR is $-0.47\,[-1.03,+0.09]$, and
$\Delta$CR is $-0.06\,[-0.65,+0.51]$ (sign-flip $p{=}0.90$). Per run, CR moves from
$8.41$ to $8.33$, $8.91$ to $8.55$, $8.33$ to $8.84$, $7.10$ to $7.17$, and $8.33$ to
$7.90$ (reference to gate, runs 42--46). A null pair with two identical arms gives
$\Delta$CR $+0.58\,[-0.65,+1.88]$. The $-0.59$ reported for AT in
Table~\ref{tab:paired_stats} from the original evaluation therefore lies within the benchmark's measurement floor.

\begin{table}[htbp]
	\centering
	\footnotesize
	\setlength{\tabcolsep}{3.5pt}
	\renewcommand{\arraystretch}{1.05}
	\begin{tabularx}{\linewidth}{@{}>{\centering\arraybackslash}Xccc>{\centering\arraybackslash}X@{}}
		\toprule
		\textbf{AT operating point} & \textbf{SR} & \textbf{FR} & \textbf{CR} & \textbf{Role} \\
		\midrule
		Planner-head PPO
			& $63.67{\scriptstyle\pm1.04}$
			& $72.48{\scriptstyle\pm0.30}$
			& $8.74{\scriptstyle\pm0.65}$
			& Ref. \\
		\trace{} calibrated gate
			& $\mathbf{64.41}$
			& $72.57$
			& $8.14$
			& Main \\
		\trace{} margin gate
			& $63.67$
			& $\mathbf{72.61}$
			& $\mathbf{8.07}$
			& FR-oriented \\
		\bottomrule
	\end{tabularx}
	\caption{\textbf{AT calibrated operating points} (drill-down of the AT
		column of Table~\ref{tab:main_compact}). Both gates were frozen before the five-run evaluation; the calibrated gate is the one reported in Table~\ref{tab:main_compact} because its selection rule (all-metric improvement) is the paper's default, and the margin gate reaches a similar CR ($8.07$) with the SR gain traded for FR. Per run the calibrated gate is not uniformly better than PPO (Table~\ref{tab:at_controls_runs}), and its paired deltas did not reproduce under concurrent evaluation (see below).}
	\label{tab:at_operating_points}
\end{table}

\paragraph{Promotion criteria.} A configuration is carried to a full five-run evaluation only if (i) on the
tuning subset, CR drops by at least $1.5$ absolute points relative to
the PPO baseline, (ii) SR does not drop by more than $0.5$ absolute
points, and (iii) the average per-step gate-fire rate stays below
$15\%$ so that the policy remains close to the base PPO behavior. The reported DT gate met all three criteria. The AT rows of Table~\ref{tab:at_operating_points} are calibrated extensions rather than the primary causal claim. Both settings carry matched slowdown, random-trigger, and lagged-risk controls (Table~\ref{tab:dt_matched_controls}, Appendix~\ref{app:at_controls}), but only on DT do all three metrics move significantly, and only the DT effect reproduces under concurrent evaluation.

\paragraph{Operating-point selection at deployment.}
For a deployment that must choose between the two AT gates, the choice reduces to
whether the user requires non-decreasing SR or non-increasing CR as the strict
constraint. The default operating point reported in Table~\ref{tab:main_compact} is
the calibrated gate because, in the original evaluation, it improved SR, FR, and CR
together. Under the concurrent re-evaluation above, neither gate changes SR, FR, or CR by a resolvable amount, so these measurements do not select between them.

\paragraph{Post-hoc $\tau$ trigger-rate curve.}
The recorded per-step risk scores on the DT gated run let us
characterize the calibration shape \emph{without} re-running eval. On
all $149{,}149$ scored steps of the full $1{,}380$-episode DT run 42 run, the
fraction of steps that would have triggered as a function of $\tau$ is:

\begin{table}[htbp]
\centering
\footnotesize
\setlength{\tabcolsep}{4pt}
\begin{tabular}{@{}cccccccccc@{}}
\toprule
$\tau$ & 0.80 & 0.84 & 0.85 & 0.86 & \textbf{0.868} & 0.88 & 0.89 & 0.90 & 0.92 \\
\midrule
step-trigger rate (\%) & 4.37 & 2.53 & 2.18 & 1.81 & \textbf{1.59} & 1.27 & 1.02 & 0.81 & 0.48 \\
\bottomrule
\end{tabular}
\caption{\textbf{Post-hoc $\tau$ trigger-rate curve on DT run 42.}
The risk-score distribution is gradual rather than bimodal at the
calibrated threshold: $\tau$ perturbations of $\pm0.02$ around $0.868$
shift trigger rate by less than $0.6$ absolute points (from $2.18\%$
at $\tau{=}0.85$ to $1.27\%$ at $\tau{=}0.88$), and even $\tau{=}0.80$, the smallest value in the table, keeps the trigger rate under $5\%$. The full
$\tau$-vs-SR/FR/CR sweep at re-evaluated trajectories is reported in
Appendix~\ref{app:tau_sensitivity}.}
\label{tab:tau_curve_posthoc}
\end{table}

\section{Auxiliary Safety Diagnostics: Time-to-Human and Min-Distance}
\label{app:tth}

\begin{table*}[htbp]
	\centering
	\footnotesize
	\setlength{\tabcolsep}{4.5pt}
	\renewcommand{\arraystretch}{1.1}
	\begin{tabular}{@{}l*{9}{c}@{}}
		\toprule
		& \multicolumn{4}{c}{\textbf{Robot--human proximity}}
		& \multicolumn{5}{c}{\textbf{Wall clearance $d_{\mathrm{obs}}$ (five runs)}} \\
		\cmidrule(lr){2-5}\cmidrule(lr){6-10}
		\textbf{Condition} & $d_t$ med. & $d_t$ p10 & TTH med. & TTH p10
		& p10 & p25 & p50 & p75 & p90 \\
		& (m) & (m) & (s) & (s) & (m) & (m) & (m) & (m) & (m) \\
		\midrule
		Planner-head PPO       & 1.520 & 1.014 & 8.558 & 3.268 & 0.027 & 0.093 & 0.221 & 0.450 & 0.715 \\
		TRACE (non-triggered)  & 1.537 & 1.038 & 8.639 & 3.320 & 0.024 & 0.086 & 0.212 & 0.437 & 0.701 \\
		TRACE (triggered)      & \textbf{0.861} & \textbf{0.622} & 11.716 & \textbf{1.278} & \textbf{0.014} & \textbf{0.043} & \textbf{0.142} & \textbf{0.254} & \textbf{0.355} \\
		\bottomrule
	\end{tabular}
	\caption{\textbf{Safety audit on DT: robot--human proximity and wall
		clearance for PPO and the \trace{} gate, split into triggered and
		non-triggered steps.} $d_t$: robot--human distance (m); TTH:
		time-to-human, $d_t/|v_x|$ (s); $d_{\mathrm{obs}}$: robot--obstacle clearance
		(m); ``med.''/p10--p90: median and percentiles (smaller $=$ closer or
		shorter). Human columns are a $600$-episode sample of the seed-42 run; wall clearance is over $230$ episodes from each of the five runs.}
\label{tab:safety_audit}
\end{table*}

To complement CR (a binary simulator contact event), we report two post-hoc safety diagnostics (Table~\ref{tab:tth_main}) on a $600$-episode sample of the \trace{} DT gated run and of the matched PPO baseline (seed 42; $59{,}343$ and $56{,}292$ steps). Every step records the robot-to-human distance $d_t$ (m) and the executed forward
velocity $v_x^{\mathrm{exec}}$ (m/s); we define
\textbf{time-to-human} $\mathrm{TTH}_t = d_t / |v_x^{\mathrm{exec}}|$
(seconds; undefined when stationary). TTH is to a human-following robot what time-to-collision (TTC) is to autonomous driving: a per-step time margin under
current velocity.

\begin{table}[htbp]
	\centering
	\footnotesize
	\setlength{\tabcolsep}{4pt}
	\begin{tabular}{@{}lccc@{}}
		\toprule
		\textbf{Metric (DT seed 42, 600-episode sample)} & \textbf{PPO} & \textbf{Gate (non-trig.)} & \textbf{Gate (trig.)} \\
		\midrule
		$n$ steps                          & 56{,}292      & 58{,}472           & 871 (1.47\%) \\
		$d_t$ median (m)                   & 1.520         & 1.537                       & 0.861 \\
		$d_t$ p10 (m)                      & 1.014         & 1.038                       & 0.622 \\
		TTH median (s)                     & 8.558         & 8.639                       & 11.716 \\
		TTH p10 (s)                        & 3.268         & 3.320                       & 1.278 \\
		\bottomrule
	\end{tabular}
	\caption{\textbf{TTH and $d_t$ on a $600$-episode sample of the matched DT seed-42 runs.}
	Triggered steps concentrate where the human is closer (median
	$d_t$ on triggered steps is $0.86$\,m vs.\ $1.54$\,m on
	non-triggered steps) and where the time margin tail is dangerous
	(TTH p10 of $1.28$\,s on triggered vs.\ $3.32$\,s on
	non-triggered). Outside the triggered window the gated run's $d_t$ medians differ from PPO's by $0.017$\,m, consistent with an intervention that is sparse and targeted rather than a blanket slowdown.}
	\label{tab:tth_main}
\end{table}

\paragraph{Min-distance per episode, grouped by outcome.}
For each episode we record the minimum $d_t$ along its trajectory.
Conditioned on \emph{collision-ending} episodes, the median min-$d_t$
is $0.455$\,m for PPO and $0.464$\,m for the gate, i.e.\ when contact occurs, both policies were equally close to the human, so the gate's
CR reduction is not coming from a different distribution of
close-approach behavior on the collided tail. Conditioned on
\emph{non-collision} episodes, the median min-$d_t$ is $1.085$\,m
(PPO) vs.\ $1.055$\,m (gate), i.e.\ the gate does not push the
follower further away from the human in nominal following. The CR
delta therefore reflects a true reduction in the number of
contact-event episodes, not a metric artifact.

\section{$\tau$ Sensitivity Sweep (SR / FR / CR)}
\label{app:tau_sensitivity}

The post-hoc trigger-rate curve (Table~\ref{tab:tau_curve_posthoc})
fixes the trajectory under $\tau{=}0.868$; the true SR/FR/CR
sensitivity to $\tau$ requires re-evaluation per setting. We sweep $\tau \in \{0.83, 0.85, 0.868, 0.88, 0.90\}$ over five DT runs (Table~\ref{tab:tau_sensitivity_full}),
keeping all other gate parameters
$(s_f, s_l, s_\omega) = (0.25, 0.60, 1.00)$ fixed. Trigger rates are
reported from the recorded seed-42 score distribution, while SR/FR/CR
are five-run means.

\begin{table}[htbp]
\centering
\footnotesize
\setlength{\tabcolsep}{4pt}
\begin{tabular}{@{}ccccc@{}}
\toprule
$\tau$ & step trigger rate (\%) & SR & FR & CR \\
\midrule
0.83 & 2.95 & 44.44 & 55.84 & 12.52 \\
0.85   & 2.18 & 44.12 & 56.24 & 12.97 \\
\textbf{0.868} (calibrated) & \textbf{1.59} & \textbf{46.02} & \textbf{57.58} & \textbf{9.83} \\
0.88   & 1.27 & 44.41 & 55.57 & 12.09 \\
0.90   & 0.81 & 44.16 & 55.87 & 12.45 \\
\bottomrule
\end{tabular}
\caption{\textbf{$\tau$ sensitivity on full DT five-run evaluation.}
The calibrated row is the main DT operating point. All four nearby thresholds keep CR below the PPO reference ($43.99/55.44/13.78$), so the safety direction does not depend on the exact threshold; the joint SR/FR/CR gain, however, is specific to the calibrated threshold, which was selected on the tuning data of Appendix~\ref{app:calibration}, and the response is not monotone in $\tau$. We report this as a limit of the operating point.}
\label{tab:tau_sensitivity_full}
\end{table}

\section{Wall Clearance in a Separate Re-evaluation}
\label{app:wall_clear}

To test whether the gate's CR reduction is a contact-event accounting artifact (i.e.\ that the gate turns ``contact'' into ``close-but-no-contact''), we recorded, at every step of a separate re-evaluation (Table~\ref{tab:wall_clearance}), the navmesh distance from the robot to the nearest obstacle, $d_{\mathrm{obs}}$ (values clipped at $2$\,m). We re-evaluated DT on a matched subset of every run (episode partitions $0$--$4$, $230$ episodes per run, $1{,}150$ per policy) for both the PPO baseline and the calibrated \trace{} gate.

\begin{table}[htbp]
\centering
\footnotesize
\setlength{\tabcolsep}{4pt}
\begin{tabular}{@{}lcccc@{}}
\toprule
\textbf{$d_{\mathrm{obs}}$ percentile (m)} & \textbf{PPO} & \textbf{Gate (all)} & \textbf{Gate (trig.)} & \textbf{Gate (non-trig.)} \\
$n$ steps & 136{,}012 & 141{,}798 & 443 & 141{,}355 \\
\midrule
p10 & 0.027 & 0.024 & 0.014 & 0.024 \\
p25 & 0.093 & 0.086 & 0.043 & 0.086 \\
p50 & 0.221 & 0.212 & 0.142 & 0.212 \\
p75 & 0.450 & 0.436 & 0.254 & 0.437 \\
p90 & 0.715 & 0.700 & 0.355 & 0.701 \\
\bottomrule
\end{tabular}
\caption{\textbf{Wall-clearance distribution under PPO vs.\ \trace{}
gate (DT, $230$ episodes from each of the five runs, $1{,}150$ per policy).} PPO and the gated run
have nearly identical per-step $d_{\mathrm{obs}}$ distributions at every
reported percentile (deltas $\leq 0.015$\,m). The gate's triggered
steps are concentrated at lower $d_{\mathrm{obs}}$ (median $0.142$\,m vs.\
$0.212$\,m non-triggered), confirming that the gate fires in close
obstacle states. \emph{Caveat}: this re-evaluation does not reproduce the headline gate's activation profile --- the gate fired on $443$ of $141{,}798$ steps ($0.31\%$, against $1.0$--$2.0\%$ in the headline runs) and collision-ending episodes moved only from $201$ to $184$ ($17.5\%$ to $16.0\%$ on this subset, against $13.78\%$ to $9.83\%$ on the full runs) --- so we use it only for the clearance distribution, which is the quantity the headline runs do not log, and not as evidence of the CR effect.}
\label{tab:wall_clearance}
\end{table}

\section{Reward Retuning and Geometric Filtering}
\label{app:negative}

Cheaper interventions were tried first, and this appendix reports the outcome of each. Table~\ref{tab:neg_compact} summarizes the
methodological alternatives that the main paper points to from
\S\ref{sec:why_not_reward}, and Figure~\ref{fig:dt_pareto_alternatives}
plots the same controls on the DT Pareto plane next to the validated
\trace{} gate. All experiments below start from the same base policy and reward. Rows marked \emph{development phase} were evaluated before the evaluator correction of Appendix~\ref{app:eval_protocol}, or are training-rollout statistics, and are therefore reported qualitatively rather than beside Table~\ref{tab:main_compact}; the distance-only filter, the geometry sweeps, the matched controls, and PPO-Lagrangian were evaluated under the corrected evaluator.

\begin{table*}[t]
	\centering
	\footnotesize
	\setlength{\tabcolsep}{4.5pt}
	\renewcommand{\arraystretch}{1.10}
	\begin{tabularx}{\textwidth}{@{}>{\raggedright\arraybackslash}p{2.7cm}L@{}}
		\toprule
		\textbf{Alternative} & \textbf{Result and takeaway} \\
		\midrule
		REINFORCE\newline (critic-free policy gradient)
		                     & \emph{Development phase (training-rollout statistics).} Collapses within the $200$-episode budget: the second round's $100$ on-policy episodes end at SR $1\%$ and CR $52\%$; not evaluated with the corrected evaluator. \emph{Takeaway:} a learned critic is necessary in this long-horizon setting. \\
		\addlinespace[2pt]
		PPO budget\newline $\times 2$
		                     & \emph{Corrected evaluator, full DT, run 42 (re-evaluated).} $46.09/56.99/16.30$ against $41.88/54.55/14.06$ for planner-head PPO under the same code ($+4.2$ SR, $+2.2$ CR): more persistence, and more contact. \emph{Takeaway:} additional RL raises success and collisions together. \\
		\addlinespace[2pt]
		Collision penalty\newline $\times 2$
		                     & \emph{Corrected evaluator, full DT, run 42 (re-evaluated).} $w_c{=}10$ with a lower learning rate and a $600$-episode budget: $41.88/55.04/21.01$ against $41.88/54.55/14.06$ for PPO under the same code, i.e.\ CR $+7$ points at equal SR. \emph{Takeaway:} scalar collision retuning produces aggressive following, not safe control. \\
		\addlinespace[2pt]
		Imitation / preference\newline (RWR, RAFT, DPO, GRPO, FTSFT)
		                     & \emph{Development phase (pre-correction evaluator).} No stable gain, or regression, under matched conditions. \emph{Takeaway:} imitation-style objectives distill rollout errors when the reference is the deployed policy. \\
		\addlinespace[2pt]
		Distance-only\newline learned filter
		                     & \emph{Corrected evaluator, full DT, run 42.} $43.84/54.70/13.55$ against PPO $44.57/56.41/12.75$. \emph{Takeaway:} distance is necessary but not sufficient as a risk feature. \\
		\addlinespace[2pt]
		Heuristic / geometry\newline filtering
		                     & \emph{Corrected evaluator, $368$-episode subset.} Navmesh-radius sweeps fail the CR-drop / SR-retention criterion. \emph{Takeaway:} safety needs richer target-state conditioning. \\ \addlinespace[2pt] PPO-Lagrangian\newline (five runs, two rates) & \emph{Corrected evaluator, full DT, five runs.} $12.70{\pm}4.25 / 32.20{\pm}5.15 / 28.67{\pm}15.33$ and $12.28{\pm}5.30 / 31.37{\pm}8.10 / 26.61{\pm}16.10$ (Table~\ref{tab:lagppo}). \emph{Takeaway:} dual-ascent retuning of the scalar objective destabilizes on every run. \\ \bottomrule
	\end{tabularx}
	\caption{\textbf{Cheaper post-training and filtering alternatives do not reproduce the safety--success pair.} All start from the same base policy and reward. Rows marked \emph{development phase} predate the evaluator correction (or are training-rollout statistics) and are qualitative; the remaining rows use the corrected evaluator. None matches \trace{}'s DT safety--success pair. Triplets read SR/FR/CR in percent. The paragraphs below expand each row.}
	\label{tab:neg_compact}
\end{table*}

\paragraph{REINFORCE.} A critic-free policy gradient using the same reward as PPO, run in the development phase. Its training rollouts collapse to a near-deterministic high-collision policy within the $200$-episode budget (second round: SR $1\%$, CR $52\%$ over $100$ on-policy episodes); the collapsed policy was not evaluated with the corrected evaluator. The collapse confirms that the learned value
head is necessary under the long horizon and noisy returns of EVT,
not just a performance optimization.

\paragraph{PPO budget $\times 2$.} A doubled rollout budget under the same hyperparameters (two further rounds from the run-42 model), re-evaluated on the full DT set under the corrected evaluator and paired with planner-head PPO under the same code (Appendix~\ref{app:learned_detector} explains the same-code pairing): SR rises from $41.88$ to $46.09$ while CR rises from $14.06$ to $16.30$, indicating that the DT problem is not a
sample-efficiency problem within this budget regime.

\paragraph{Collision penalty $\times 2$.} Doubling $w_c$ to $10$, in a configuration that also used a lower learning rate and a $600$-episode budget, re-evaluated on the full DT set under the corrected evaluator (run 42): $41.88/55.04/21.01$ against $41.88/54.55/14.06$ for PPO under the same code. The policy becomes more aggressive in following without becoming safer, and DT CR worsens by $7$ points. This behavior is consistent
with the marginal shift of
Figure~\ref{fig:main_dt_evidence}a--b:
stronger following pressure moves more endings from lost-after-following
toward collision-while-following.

\paragraph{Task-specific PPO.} Two development-phase variants train on DT-only or transfer from STT to DT (pre-correction evaluator). Neither beats the STT-trained planner-head PPO model on DT, ruling out a simple
training-task mismatch.

\paragraph{Imitation / preference objectives.} RWR variants, self-SFT/FTSFT, RAFT, DPO, and GRPO, run in the development phase under the pre-correction evaluator, all show no stable gain or regress under matched conditions. We attribute this to the
fact that imitation-style objectives distill rollout errors when the
supervised reference is the deployed policy itself.

\paragraph{Recognition refresh SFT.} A small-data SFT on the recognition pathway, evaluated in the development phase on a $368$-episode subset with the pre-correction evaluator. DT refresh SFT and
recognition-only refresh fail the promotion criteria. PPO-initialized clean
refresh is less destructive but still below planner-head PPO.

\paragraph{Geometry / distance-only filtering.} Navmesh radius sweeps ($368$-episode subset) and a learned distance-only filter (full DT, run 42), both under the corrected evaluator. None pass
the CR-drop / SR-retention criterion. The learned distance-only filter underperforms the PPO seed-$42$ reference, showing that distance is
necessary but not sufficient as a risk feature.

\paragraph{Matched action-scaling controls.}
The three controls---average slowdown, random-trigger
slowdown, and lagged-risk slowdown---are discussed in
\S\ref{sec:why_not_reward} and Table~\ref{tab:dt_matched_controls}, with per-run numbers in Table~\ref{tab:dt_matched_controls_per_seed}. All fifteen full-DT evaluations ($3$ controls $\times$ $5$ runs, $30$ episode partitions and $1{,}380$ episodes per run/control) reuse the same risk head, threshold, scales, target-state features, and probe as the headline \trace{} run, so the trigger source is the intended difference; trigger rates are matched approximately rather than exactly: the gate fires on $1.59/1.69/1.95/1.02/1.88\%$ of steps on runs $42$--$46$ ($1.6\%$ pooled; $31.5/32.6/34.1/24.2/32.1\%$ of episodes), random on $1.1$--$2.0\%$ and lagged-risk on $0.5$--$0.9\%$ (Table~\ref{tab:dt_matched_controls_per_seed}). Average slowdown applies the per-run
trigger-rate-implied mean attenuation at every step; random slowdown samples timesteps
approximately at each run's measured gate rate; lagged-risk uses the gate's risk head and threshold but reads a $37$-step lagged score. None reproduces the gate's safety--success pair. These controls test whether the gate's small attenuation budget is sufficient on its own; they do not test a strong uniform slowdown or a hand-crafted proximity trigger at a matched rate, which would probe whether the learned head is necessary.

\begin{table*}[t]
	\centering
	\scriptsize
	\setlength{\tabcolsep}{3.2pt}
	\renewcommand{\arraystretch}{1.03}
	\begin{tabular}{llrrrrrrr}
		\toprule
		\textbf{Control} & \textbf{Run} & \textbf{SR}                & \textbf{FR}        & \textbf{CR} &
		\textbf{$\Delta$SR vs PPO} & \textbf{$\Delta$FR vs PPO} & \textbf{$\Delta$CR vs PPO} & \textbf{Step rate}                                                  \\
		\midrule
		average                    & 42                         & 42.75                      & 55.01              & 13.84       & -1.82 & -1.40 & +1.09 & 100.00\% \\
		average                    & 43                         & 43.26                      & 55.10              & 15.29       & +0.00 & +0.49 & +0.07 & 100.00\% \\
		average                    & 44                         & 43.55                      & 55.83              & 13.77       & -0.73 & +0.19 & -1.45 & 100.00\% \\
		average                    & 45                         & 44.78                      & 55.53              & 11.96       & +0.72 & +0.28 & -0.79 & 100.00\% \\
		average                    & 46                         & 42.61                      & 54.21              & 13.48       & -1.16 & -1.07 & +0.51 & 100.00\% \\
		random                     & 42                         & 42.83                      & 55.04              & 14.06       & -1.74 & -1.37 & +1.31 & 1.36\%   \\
		random                     & 43                         & 44.28                      & 55.43              & 14.57       & +1.02 & +0.82 & -0.65 & 1.81\%   \\
		random                     & 44                         & 43.55                      & 55.11              & 14.57       & -0.73 & -0.53 & -0.65 & 1.99\%   \\
		random                     & 45                         & 44.49                      & 56.10              & 12.54       & +0.43 & +0.85 & -0.21 & 1.13\%   \\
		random                     & 46                         & 44.64                      & 55.66              & 12.61       & +0.87 & +0.38 & -0.36 & 1.66\%   \\
		lagged                   & 42                         & 42.54                      & 54.72              & 13.48       & -2.03 & -1.69 & +0.73 & 0.81\%   \\
		lagged                   & 43                         & 42.90                      & 55.84              & 14.93       & -0.36 & +1.23 & -0.29 & 0.66\%   \\
		lagged                   & 44                         & 43.48                      & 55.24              & 14.71       & -0.80 & -0.40 & -0.51 & 0.90\%   \\
		lagged                   & 45                         & 44.71                      & 55.89              & 12.61       & +0.65 & +0.64 & -0.14 & 0.45\%   \\
		lagged                   & 46                         & 44.64                      & 56.18              & 14.28       & +0.87 & +0.90 & +1.31 & 0.94\%   \\
		\bottomrule
	\end{tabular}
	\caption{\textbf{Full matched controls, per run.} All rows use full
		DT evaluation with $1{,}380$ episodes. Positive $\Delta$CR is worse. The
		controls stay close to PPO and remain far from \trace{}'s
		$46.02$ / $57.58$ / $9.83$ five-run mean, supporting the conclusion
		that the gate's gain is not explained by slowdown, sparse trigger
		budget, or risk-score density alone.}
	\label{tab:dt_matched_controls_per_seed}
\end{table*}

\paragraph{PPO-Lagrangian (five runs, two dual-ascent rates).}
A principled constrained-RL alternative keeps the scalar objective and adds a Lagrangian collision constraint: the collision indicator of Equation~\ref{eq:reward} is weighted by an additional multiplier $\lambda$, updated once per round by dual ascent, $\lambda \leftarrow \max(0, \lambda + \eta\,(\mathrm{CR}_{\mathrm{round}} - 0.10))$, with $\lambda_0 = 5$ and $\eta \in \{20, 5\}$, over four rounds of $50$ on-policy episodes (the same $200$-episode budget as planner-head PPO's two rounds of $100$) from the same base. On run 42 with $\eta{=}20$ the multiplier moves $5 \to 13.8 \to 11.8 \to 9.8 \to 7.8$ while the rollouts collapse to zero following after the first round, so the failure is a collapse under the constraint rather than a slow drift; we did not tune the dual-ascent schedule beyond the two rates. It destabilizes on every run. With a dual-ascent rate of $20$ the five
runs reach $12.70{\pm}4.25$ / $32.20{\pm}5.15$ / $28.67{\pm}15.33$ (SR / FR / CR), and a softer rate of $5$ collapses the same
way ($12.28{\pm}5.30$ / $31.37{\pm}8.10$ / $26.61{\pm}16.10$), so the collapse is not specific to one dual-ascent rate at this budget (Table~\ref{tab:lagppo}). The five planner-head PPO runs it is compared with reach
$43.99$ / $55.44$ / $13.78$.

\begin{table}[htbp]
	\centering
	\footnotesize
	\setlength{\tabcolsep}{6pt}
	\begin{tabular}{lcc}
		\toprule
		\textbf{Run} & \textbf{$\lambda$-rate $20$: SR / FR / CR} & \textbf{$\lambda$-rate $5$: SR / FR / CR} \\
		\midrule
		42 & $12.83$ / $33.37$ / $26.45$ & $12.97$ / $34.77$ / $10.43$ \\
		43 & $11.88$ / $34.74$ / $14.28$ & $14.71$ / $37.33$ / $14.78$ \\
		44 & $19.57$ / $37.84$ / $14.71$ & $18.99$ / $38.13$ / $37.25$ \\
		45 & $7.97$ / $24.16$ / $49.71$ & $9.86$ / $27.77$ / $21.67$ \\
		46 & $11.23$ / $30.91$ / $38.19$ & $4.86$ / $18.85$ / $48.91$ \\
		\midrule
		mean$\pm$std & $12.70{\pm}4.25$ / $32.20{\pm}5.15$ / $28.67{\pm}15.33$ & $12.28{\pm}5.30$ / $31.37{\pm}8.10$ / $26.61{\pm}16.10$ \\
		\bottomrule
	\end{tabular}
	\caption{\textbf{PPO-Lagrangian on DT, per run.} Same base, reward, rollout budget,
		and corrected evaluator as planner-head PPO; only the constraint machinery is
		added. Both dual-ascent rates destabilize training on every run.}
	\label{tab:lagppo}
\end{table}

\section{Single-Target Tracking: Alternatives Considered}
\label{app:stt_attempts}

Planner-head PPO on single-target tracking raises SR by $8.2$ points
and FR by $5.2$ points over the supervised baseline while leaving CR
essentially unchanged ($4.13 \rightarrow 4.09$). Any further extension
must meet a strict promotion criterion---non-degrading CR with at least
$+0.25$ SR and $+0.25$ FR on an independent validation subset---rather
than merely improve tracking. Table~\ref{tab:stt_attempts} reports the completed alternatives; micro-gate and gated planner apply the DT gate to the STT planner at two operating points, collision-anchored continuation is one further PPO round anchored on the collision term, the two gate-repair rows re-tune the gate on top of it (on all runs, and on the two weakest runs), and the last three rows are the $c{=}16$ and $c{=}18$ continuations and their per-run mix. Plain planner continuation can raise
SR and FR but does not by itself meet the CR criterion: the best
collision-anchored continuation remains $+0.05$ above PPO on CR.
Every gate-repair configuration we tried produced
a trigger rate of exactly $0\%$ on both episodes and steps, so the
runtime-gate interface cannot supply the missing safety lever on
this setting. The two single-$c$ planner continuations ($c{=}16$ and
$c{=}18$) each land just above the PPO CR in full five-run mean,
but inspection of the selection and independent subsets shows they
have \emph{complementary} per-run CR profiles: $c{=}16$ is CR-fail
on the selection subset and strict-Pareto on the independent subset; $c{=}18$ is
the reverse. The per-run CR-anchored selection chooses, for each run, the collision-penalty level
	with lower selection-subset CR (selection rule:
	$\{42{:}c18, 43{:}c16, 44{:}c18, 45{:}c18, 46{:}c16\}$); the resulting
mix is strict-Pareto on all three subsets (selection, independent,
full five-run) and is the row reported in Table~\ref{tab:main_compact}.

\begin{table}[t]
	\centering
	\footnotesize
	\setlength{\tabcolsep}{3pt}
	\renewcommand{\arraystretch}{1.04}
	\begin{tabularx}{\textwidth}{@{}>{\raggedright\arraybackslash}p{2.65cm}>{\raggedright\arraybackslash}p{1.55cm}>{\centering\arraybackslash}p{2.75cm}X@{}}
		\toprule
		\textbf{Attempt} & \textbf{Scope} & \textbf{SR / FR / CR} & \textbf{Decision} \\
		\midrule
		Planner-head PPO & 5 runs & $86.45 / 80.88 / 4.09$ & Main STT reference. \\
		Micro-gate on planner & 5 runs & $86.39 / 80.73 / 4.06$ & Essentially neutral; no tracking gain. \\
		Gated planner & 5 runs & $86.84 / 81.27 / 4.30$ & SR and FR improve, but CR worsens. \\
		Collision-anchored continuation & 5 runs & $87.35 / 81.69 / 4.14$ & Strongest tracking among single continuations; CR remains $0.05$ above PPO. \\
			Gate-repair sweep & selection subset & $88.84 / 84.35 / 2.61$ & Best selection-subset row among four gate settings, all of which trigger on $0\%$ of episodes and steps, so the shift comes from the planner. \\
		Weak-run gate repair & runs 44/45 & not aggregated & CR rises on both diagnostic runs while the gate still triggers on $0\%$ of episodes and steps. \\
			CR-constrained selection sweep & selection subset & $90.29 / 85.41 / 2.32$ & Three candidates meet the selection criterion after one extra low-learning-rate PPO round; two are carried to full five-run evaluation. \\
		CR-preserving continuation ($c{=}16$) & 5 runs & $87.26 / 81.68 / 4.13$ & Improves SR/FR over PPO but leaves CR $0.04$ above PPO; strict-Pareto only on the independent subset. \\
		Balanced continuation ($c{=}18$) & 5 runs & $87.36 / 81.77 / 4.14$ & Repairs runs $44$ and $45$ but inflates run $46$; strict-Pareto only on the selection subset. \\
		\textbf{Per-run CR-anchored mix} & 5 runs + independent & $\mathbf{87.51 / 81.80 / 3.91}$ & \textbf{Reported in Table~\ref{tab:main_compact}.} Uses selection-subset CR to select $c{=}16$ or $c{=}18$ per run; strict-Pareto on all three subsets. \\
		\bottomrule
	\end{tabularx}
	\caption{\textbf{Completed STT attempts.} Values are SR/FR/CR. The two
		single-$c$ continuations are each strict-Pareto on exactly one evaluation
		subset; the per-run CR-anchored mix selects the lower-CR variant per run using only the selection subset and remains strict-Pareto on selection,
		independent, and full five-run settings. Gate-repair diagnostics are retained
		because they show that the runtime gate is not the active lever on STT.}
	\label{tab:stt_attempts}
\end{table}

\paragraph{Selection-grid stability.}
To test whether the two-level mix depends on its grid, we trained three further
penalty levels per run ($c\in\{15,17,20\}$; $15$ trainings and $15$ full evaluations) and
re-ran the identical a priori rule --- minimum selection-subset CR, ties to the lower
penalty --- on the five-level grid. The selection subset is the first $690$ episodes of each run in evaluation order (partitions $0$--$14$) and the independent subset is the remaining $690$; the two sets coincide on four runs and differ by $25$ episodes on run 43. On the selection subset adjacent levels differ by at most five episodes ($0.72$ points), and the two picks that change do so at an exact tie (run 42, four levels at $2.46$) or a three-episode gap (run 43, $2.32$ against $2.75$), so the rule is choosing among near-equivalent models (Table~\ref{tab:stt_grid}). The re-selected mix reaches $86.00 / 78.14 / 4.67$ on the independent subset against $85.77 / 78.05 / 4.81$ for the two-level mix of Table~\ref{tab:main_compact} and $84.99 / 77.11 / 5.01$ for PPO on the same subset (its full-set values are in Table~\ref{tab:main_compact}): both mixes are strict-Pareto over PPO on this subset, as are three of the five fixed levels ($c{=}15$: $86.09/78.10/4.99$; $c{=}16$: $85.74/77.91/4.84$; $c{=}20$: $85.74/77.88/4.99$; $c{=}17$ at $85.62/77.82/5.01$ and $c{=}18$ at $85.65/77.90/5.07$ are not). The difference is that a fixed level can only be chosen by looking at held-out data, whereas the rule's pick is made on the selection subset alone: the two-level mix remains strict-Pareto on the selection, independent, and full sets (Table~\ref{tab:stt_attempts}), and the five-level mix on the independent and full sets ($87.58 / 81.89 / 3.80$ against PPO's $86.45 / 80.88 / 4.09$ on the full set). The result therefore does not depend on which grid the rule is offered.

\begin{table}[htbp]
	\centering
	\footnotesize
	\setlength{\tabcolsep}{5pt}
	\begin{tabular}{lcccccll}
		\toprule
		\textbf{Run} & $c{=}15$ & $c{=}16$ & $c{=}17$ & $c{=}18$ & $c{=}20$ & \textbf{Two-level pick} & \textbf{Five-level pick} \\
		\midrule
		42 & $\mathbf{2.46}$ & $3.04$ & $\mathbf{2.46}$ & $\mathbf{2.46}$ & $\mathbf{2.46}$ & $c{=}18$ & $c{=}15$ \\
		43 & $\mathbf{2.32}$ & $2.75$ & $2.75$ & $3.04$ & $2.46$ & $c{=}16$ & $c{=}15$ \\
		44 & $2.90$ & $3.48$ & $3.04$ & $\mathbf{2.75}$ & $3.04$ & $c{=}18$ & $c{=}18$ \\
		45 & $3.91$ & $4.49$ & $3.91$ & $\mathbf{3.77}$ & $3.91$ & $c{=}18$ & $c{=}18$ \\
		46 & $4.06$ & $\mathbf{3.33}$ & $\mathbf{3.33}$ & $4.06$ & $3.48$ & $c{=}16$ & $c{=}16$ \\
		\bottomrule
	\end{tabular}
	\caption{\textbf{Selection-subset CR (\%) per run and penalty level.} Bold marks the minimum on the five-level grid. The two picks that change (runs 42 and 43) do so at a tie or a three-episode gap; the two-level rule is the one used for Table~\ref{tab:main_compact}.}
	\label{tab:stt_grid}
\end{table}

\section{Episode-Ending Taxonomy and Transition Analysis}
\label{app:identity}

\begin{figure}[t]
	\centering
	\includegraphics[width=\linewidth]{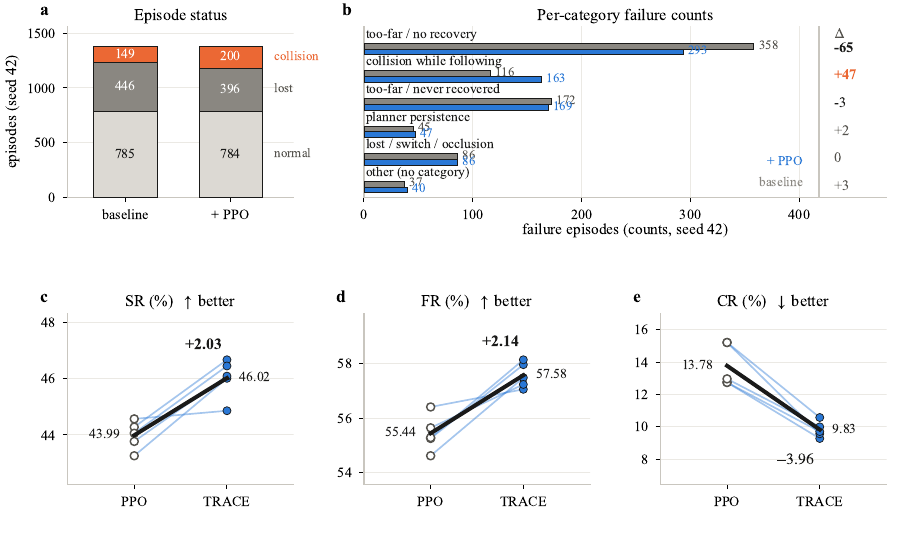}
	\caption{\textbf{DT ending-taxonomy evidence and five-run gate effect.}
		\textbf{a--b}, On one matched diagnostic run (seed 42) planner-head PPO leaves the number of normal endings essentially unchanged ($785$ to $784$) while shifting the ending distribution from
		lost toward collision: too-far/no-recovery failures fall by $65$ and
		collision-while-following failures rise by $47$ (Table~\ref{tab:dt_taxonomy_app});
		the episode-level transitions (Table~\ref{tab:dt_transition}) show a net shift, not
		a one-to-one conversion.
		\textbf{c--e}, PPO$\rightarrow$\trace{} on all five independent training runs,
		aggregated by the official evaluator --- the same convention as
		Table~\ref{tab:main_compact}, so the panel means reproduce its paired row
		exactly ($+2.03$/$+2.14$/$-3.96$). Hollow $=$ PPO, filled $=$ \trace{}, thin
		lines pair the runs, the black line joins the means. All three metrics improve
		on all five runs; CR falls by $2.97$ to $5.65$ points.}
	\label{fig:main_dt_evidence}
\end{figure}

\paragraph{Definition of the proxy.}
The diagnostic evaluator records, per step, a detector-to-panoptic
overlap proxy for the selected target. The proxy is the IoU between
the detector box of the policy's selected target and the panoptic
instance mask of the ground-truth target. A successful identity is one
where the proxy stays above a small threshold ($\geq 0.20$ in the
released analysis) across most of the followed timesteps.

\paragraph{Failure taxonomy.}
We classify each non-success episode into one of five categories:
too-far or no-recovery after following; too-far or never recovered;
collision while following; lost after following with switch or
occlusion; planner persistence or late failure. The categories are mutually
exclusive but not exhaustive: $37$ ($4.5\%$) baseline and $40$ ($5.0\%$) PPO non-success
episodes match none of them and are reported as \emph{other}; see
Figure~\ref{fig:main_dt_evidence}b and Table~\ref{tab:dt_taxonomy_app}.

\begin{table}[ht]
	\centering
	\footnotesize
	\setlength{\tabcolsep}{4pt}
	\resizebox{\textwidth}{!}{%
		\begin{tabular}{@{}lccc@{}}
			\toprule
			\textbf{DT ending diagnostic}           & \textbf{Compact baseline} & \textbf{+ Planner-head PPO (run 42)} & \textbf{Change}       \\
			\midrule
			SR / FR / CR                                    & 41.01 / 53.37 / 10.80     & 42.17 / 54.45 / 14.49             & +1.16 / +1.08 / +3.69 \\
			Episode status: normal / lost / collision       & 785 / 446 / 149           & 784 / 396 / 200                   & $-1$ / $-50$ / +51    \\
			\midrule
			too-far / no-recovery after following           & 44.0\% (358)              & 36.7\% (293)                      & $-65$ failures        \\
			too-far / never recovered                       & 21.1\% (172)              & 21.2\% (169)                      & $-3$ failures         \\
			collision while following                       & 14.3\% (116)              & 20.4\% (163)                      & $+47$ failures        \\
			lost after following / switch / occlusion proxy & 10.6\% (86)               & 10.8\% (86)                       & unchanged             \\
			planner persistence or late failure proxy       & 5.5\% (45)                & 5.9\% (47)                        & $+2$ failures         \\
			other (matches no category)                     & 4.5\% (37)                & 5.0\% (40)                        & $+3$ failures         \\
			\bottomrule
		\end{tabular}}
	\caption{\textbf{Full DT identity / failure-case diagnostic
			(single matched seed).} Numeric backing for
		Figure~\ref{fig:main_dt_evidence}a--b. The SR/FR/CR triplets differ from the headline run-42 evaluation ($44.57/56.41/12.75$ for PPO) because the diagnostic evaluator is a separate pass whose per-step logging changes outcomes by up to $2.4$ SR and $1.7$ CR points; we use this run only for the failure taxonomy, never as a performance estimate.
		The per-category counts are from one matched seed; the aggregate
		five-run CR rise of $+2.77$ in Table~\ref{tab:main_compact} is
		consistent with the same shift.}
	\label{tab:dt_taxonomy_app}
\end{table}

\paragraph{Caveats.}
The proxy is detector-conditioned, so detector failure modes (missed
detections, identity switches between detector tracklets) propagate
into the proxy. We use the proxy only to compare distributions of
failure categories under matched evaluators, not to claim absolute
target-switch rates.

\paragraph{Multi-run status mix: the gate reverses the shift on every run.}
Table~\ref{tab:dt_taxonomy_app} reports the baseline$\rightarrow$PPO
direction on a single matched seed (run 42). Here we extend the diagnosis
to the PPO$\rightarrow$gate direction on \emph{all five} matched
DT runs, using the same full $1{,}380$-episode protocol (Table~\ref{tab:dt_taxonomy_multiseed}):

\begin{table}[htbp]
	\centering
	\footnotesize
	\setlength{\tabcolsep}{4pt}
	\begin{tabular}{@{}crrrrrr@{}}
			\toprule
			\textbf{Run} & \textbf{$\Delta$SR} & \textbf{$\Delta$FR} & \textbf{$\Delta$CR} &
			\textbf{$\Delta$Normal} & \textbf{$\Delta$Lost} & \textbf{$\Delta$Collision} \\
			\midrule
			42 & $+0.29$ & $+0.65$ & $\mathbf{-3.48}$ & $+22$ & $+26$ & $\mathbf{-48}$ \\
			43 & $+2.75$ & $+2.88$ & $\mathbf{-4.64}$ & $+57$ & $+7$  & $\mathbf{-64}$ \\
			44 & $+2.39$ & $+1.60$ & $\mathbf{-5.65}$ & $+45$ & $+33$ & $\mathbf{-78}$ \\
			45 & $+2.39$ & $+2.72$ & $\mathbf{-3.04}$ & $+48$ & $-6$  & $\mathbf{-42}$ \\
			46 & $+2.32$ & $+2.87$ & $\mathbf{-2.97}$ & $+34$ & $+7$  & $\mathbf{-41}$ \\
			\midrule
			Mean & $\mathbf{+2.03}$ & $\mathbf{+2.14}$ & $\mathbf{-3.96}$ & $\mathbf{+41.2}$ & $\mathbf{+13.4}$ & $\mathbf{-54.6}$ \\
		\bottomrule
	\end{tabular}
	\caption{\textbf{PPO$\rightarrow$gate deltas per matched DT run ($1{,}380$ episodes each).} SR, FR and CR are aggregated by the official
		EVT-Bench aggregator --- the same convention as Table~\ref{tab:main_compact},
		so the run means here reproduce its paired row exactly ($+2.03$/$+2.14$/$-3.96$).
		All three metrics improve on all five runs; endings shift from collision mostly to normal ($+41.2$ per run) and less to lost ($+13.4$), so the collision reduction is not purchased mainly by abandoning the target. Episode-level intervals for these deltas are in Table~\ref{tab:paired_stats}; we prefer them to the five-run one-sided sign test ($p{=}0.031$), which this table would also pass. Per-run exact McNemar $p$ on collisions: DT below $2{\times}10^{-4}$ on every run; AT $0.15/0.022/1.0/0.91/1.0$; STT $0.33/0.23/0.68/1.0/0.15$.}
	\label{tab:dt_taxonomy_multiseed}
\end{table}

\section{Recovery Residual Dataset Construction}
\label{app:residual_data}

\paragraph{Source.}
Source episodes come from a full DT evaluation run with the \trace{} gate enabled and identity debugging on. The run reports $1{,}380$
episodes at $45.29$ / $56.55$ / $9.64$ (SR / FR / CR) and provides target-state, simulator-pose, and recovery-event annotations per step.

\paragraph{Label windows.}
For each lost or too-far event, we mark a positive window of width $K$
steps preceding the event. Three variants are produced:
(i) the \emph{default} variant with the longest positive window, yielding $11{,}214$
rows at $56.14\%$ positives;
(ii) a \emph{short-20} variant narrows the window and yields $23{,}831$ rows at
$20.07\%$ positives;
(iii) a \emph{short-25} variant is intermediate at $19{,}232$ rows and $28.22\%$
positives. Negative rows are sampled uniformly from non-event timesteps
in the same episodes.

\paragraph{Targets.}
Each row carries a ground-truth delta
$(\Delta v_x, \Delta v_y, \Delta \omega)$ toward the target in the
corrected robot frame ($+X$ forward). Rows without a valid robot-frame pose are dropped at construction time; the released datasets
contain zero missing-pose rows.

\paragraph{Partitions.}
Each variant is partitioned episode-disjointly into train / val / test
partitions ($0.8$ / $0.1$ / $0.1$) so that supervised checks are evaluated on episodes unseen during head training.

\section{Robot-Frame Grounding Pipeline}
\label{app:robotframe}

\paragraph{Ground-plane convention.} The initial target-grounding data collectors used local $-Z$ as the
ground-plane forward axis. This axis is close to vertical for the
embodied agent we study, while the executed base velocity uses local
$+X$. Bearings computed against $-Z$ are therefore systematically
rotated by approximately ninety degrees relative to the controller's
actual forward direction.

\paragraph{Corrected convention.} The corrected collector projects target position into the local
ground-plane frame using local $+X$ as forward and local $+Y$ as left.
Bearing is the signed angle between the projected target vector and
the forward axis. The same convention is used to label
left / front / right buckets in the failure taxonomy.

\begin{figure*}[!htbp]
	\centering
	\begin{minipage}[t]{0.49\textwidth}
		\centering
		\includegraphics[width=\linewidth]{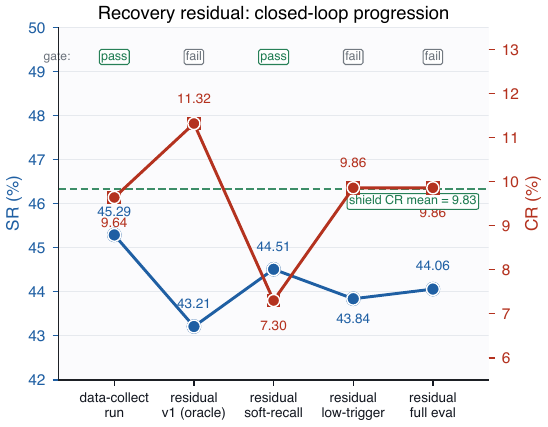}
	\end{minipage}\hfill
	\begin{minipage}[t]{0.49\textwidth}
		\centering
		\includegraphics[width=\linewidth]{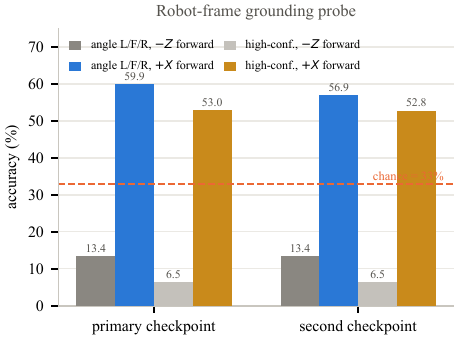}
	\end{minipage}
	\caption{\textbf{Recovery-residual progression and robot-frame
			grounding probe.} \textbf{(a)} Supervised recovery residuals,
		closed-loop progression: SR (blue, left axis) and CR (red, right
		axis) for the data-collection run and four residual
		configurations. Dashed green line marks the \trace{} gate CR
		(five-run mean $9.83$); soft-recall meets the validation promotion criterion (CR $7.3$) but its full DT seed-$42$ evaluation rises back above
		the gate, motivating a diagnosis based on the target's state.
		\textbf{(b)} The robot-frame grounding correction recovers the deployable direction probe: validation angle L/F/R and high-confidence-subset
		accuracy before (grey, $-Z$ forward) and after (blue / ochre, $+X$ forward) the convention correction, on two trained models; dashed
		line is $33\%$ chance.}
	\label{fig:recovery_grounding}
\end{figure*}

\paragraph{Effect on the deployable probe (offline) and verification.}
We retrain the deployable target-direction probe on the corrected
labels. Under the $-Z$ convention the probe was below chance ($13.4\%$ angle L/F/R,
$6.5\%$ high-confidence), making it unusable as a closed-loop signal.
Under the corrected $+X$ convention it reaches $59.9\%$ / $53.0\%$ on the primary model and $56.9\%$ / $52.8\%$ on a second model
(Figure~\ref{fig:recovery_grounding}b), comfortably above chance on the
high-confidence subset---the regime relevant for recovery control.
Closed-loop verification follows a two-gate protocol (a probe criterion for the direction signal alone, a recovery criterion for the controller on top); a configuration is carried forward only after meeting both on a second model, so any
future recovery claim is attributable to the probe correction
\emph{or} the controller, never to a mixture.

\end{document}